\documentclass[]{mosi}

\usepackage{amsmath} 
\usepackage{natbib}
\usepackage{graphicx}
\usepackage{subcaption} 

\usepackage[toc,page,header]{appendix}
\usepackage[T1]{fontenc}    
\usepackage{hyperref}       
\usepackage{url}            
\usepackage{booktabs}       
\usepackage{amsfonts}       
\usepackage{nicefrac}       
\usepackage{microtype}      
\usepackage{wrapfig}

\usepackage{amssymb}        
\usepackage{titletoc}
\usepackage{minitoc}

\usepackage{array}
\usepackage{etoolbox}

\definecolor{lightblue}{RGB}{200, 230, 255}  
\definecolor{headerblue}{RGB}{150, 200, 255} 

\usepackage{pgfplots}
\usepackage{xcolor}
\usepackage{float}
\usepackage{comment}
\usepackage{multirow}
\usepackage{makecell}
\usepackage{siunitx}
\usepackage{tikz}
\usepackage{pgf-pie}
\usepackage[export]{adjustbox}

\usepackage{ragged2e}
\usepackage{tabularx}
\usepackage{caption}
\usepackage{enumitem}
\usepackage{pifont}
\usepackage[hang,flushmargin]{footmisc}

\usepackage{tcolorbox}
\tcbuselibrary{breakable}
\tcbuselibrary{skins}

\usepackage{tabularx}
\usepackage{listings}

\usepackage{inconsolata}

\usepackage{algorithm}
\usepackage{algorithmic}
\usepackage{adjustbox}
\usepackage{multirow}
\usepackage{tcolorbox}
\usepackage[table]{xcolor}
\usepackage{nicematrix}
\usepackage{booktabs} 
\usepackage{amsmath}
\usepackage{float}
\usepackage{xspace}
\usepackage{subcaption}
\usepackage{calc} 
\usepackage{cleveref}

\usepackage{threeparttable}
\usepackage{setspace}
\usepackage[scheme=plain]{ctex} 

\crefname{figure}{Fig.}{Figs.}
\crefname{table}{Tab.}{Tabs.}
\crefname{section}{Sec.}{Secs.}
\crefname{equation}{Eq.}{Eqs.}
\crefname{appendix}{App.}{Apps.}

\definecolor{MossCyan}{HTML}{82D9FF} 
\definecolor{MossBlue}{HTML}{82B1FF}

\definecolor{ForestGreen}{RGB}{34, 139, 34}
\definecolor{Red}{RGB}{255, 0, 0}

\definecolor{tickG}{rgb}{0.1, 0.588, 0.1}
\definecolor{crossR}{rgb}{0.588, 0.1, 0.1}

\newcommand{\titleLogo}{%
    $\raisebox{-1.5mm}
    {\includegraphics[width=0.05\textwidth]{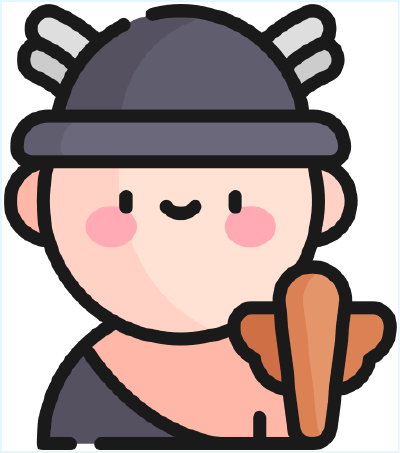}}$ 
}

\definecolor{frenchblue}{rgb}{0.0, 0.45, 0.73}
\definecolor{babyblue}{rgb}{0.54, 0.81, 0.94}
\definecolor{classicrose}{rgb}{0.98, 0.8, 0.91}
\definecolor{beige}{rgb}{0.96, 0.96, 0.86}
\definecolor{forestgreen}{HTML}{2e7d43}

\definecolor{blue1}{HTML}{91BBE6}
\definecolor{blue2}{HTML}{3F90E0}
\definecolor{blue3}{HTML}{316FAD}

\definecolor{color1}{HTML}{FF9999}
\definecolor{color2}{HTML}{FF6666}
\definecolor{color3}{HTML}{FF3333}
\definecolor{color4}{HTML}{E60000}
\definecolor{color5}{HTML}{B30000}
\definecolor{color6}{HTML}{8CD98C}
\definecolor{color7}{HTML}{53c653}
\definecolor{color8}{HTML}{00B050}
\definecolor{color9}{HTML}{2d862d}
\definecolor{color10}{HTML}{206020}
\definecolor{color11}{HTML}{cca300}
\definecolor{HermesBlue}{RGB}{0,102,204}

\newcommand{\hermes}{\textcolor{HermesBlue}{\textit{\textbf{HERMES}}}\xspace}

\newtcolorbox{promptbox}[2][]{
    colback=white,
    coltext=black,
    arc=3mm,
    boxrule=0.5pt,
    colframe=black!60!white,
    title={#2},
    colbacktitle=black,
    coltitle=white,
    fonttitle=\bfseries,
    top=8pt,
    bottom=8pt,
    left=10pt,
    right=10pt,
    breakable,
    before upper={%
        \linespread{1}\selectfont
        \setlength{\parskip}{1ex plus 0.2ex minus 0.2ex}%
        \setlength{\parindent}{0pt}%
    },
    #1
}

\title{\titleLogo \hermes: KV Cache as Hierarchical Memory for Efficient Streaming Video Understanding}

\author{
Haowei Zhang$^{1,*}$\hspace{.1em}
Shudong Yang$^{1,2,*}$\hspace{.1em}
Jinlan Fu$^{1,\dagger}$ \hspace{.1em}
See-Kiong Ng$^{3}$ \hspace{.1em}
Xipeng Qiu$^{1,2,\dagger}$ \hspace{.1em}
}
\affiliation{
  $^1$Fudan University, $^2$Shanghai Innovation Institute,
  $^3$National University of Singapore
}

\abstract{
\begin{abstract}
\noindent Recent advancements in Multimodal Large Language Models (MLLMs) have demonstrated significant improvement in offline video understanding. However, extending these capabilities to streaming video inputs, remains challenging, as existing models struggle to simultaneously maintain stable understanding performance, real-time responses, and low GPU memory overhead. To address this challenge, we propose \hermes, a novel training-free architecture for real-time and accurate understanding of video streams. Based on a mechanistic attention investigation, we conceptualize KV cache as a hierarchical memory framework that encapsulates video information across multiple granularities. During inference, \hermes reuses a compact KV cache, enabling efficient streaming understanding under resource constraints. Notably, \hermes requires no auxiliary computations upon the arrival of user queries, thereby guaranteeing real-time responses for continuous video stream interactions, which achieves 10$\times$ faster TTFT compared to prior SOTA. Even when reducing video tokens by up to 68\% compared with uniform sampling, \hermes achieves superior or comparable accuracy across all benchmarks, with up to 11.4\% gains on streaming datasets.
\end{abstract}
}

\checkdata[Correspondence]{\email{hwzhang25@m.fudan.edu.cn}, \email{jinlanjonna@gmail.com}, \email{xpqiu@fudan.edu.cn}}
\checkdata[Homepage]{\url{https://hermes-streaming.github.io/}}
\checkdata[Repository]{\url{https://github.com/haowei-freesky/HERMES}}

\newcommand{\llava}{LLaVA-OV-7B\xspace}
\newcommand{\llavasmall}{LLaVA-OV-0.5B\xspace}
\newcommand{\qwen}{Qwen2.5-VL-7B\xspace}

\definecolor{grey}{RGB}{128,138,135}

\begin{document}
\maketitle
\begingroup
\renewcommand{\thefootnote}{\fnsymbol{footnote}}
\setcounter{footnote}{1}
\footnotetext{Equal contribution.}
\setcounter{footnote}{2}
\footnotetext{Corresponding author.}
\endgroup



\newlength{\myfigheight}
\setlength{\myfigheight}{7.01cm} 

\begin{figure*}[t]
    \centering
    
    \begin{subfigure}{\widthof{\includegraphics[height=\myfigheight, trim={0 0 415pt 0}, clip]{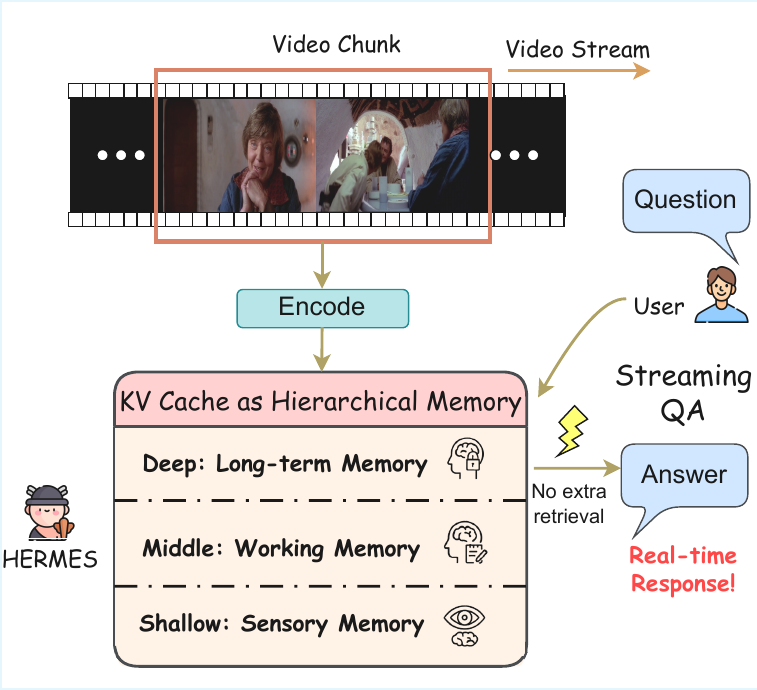}}}
        \centering
        \includegraphics[height=\myfigheight, trim={0 0 415pt 0}, clip]{figures/teaser.pdf}
        \caption{\hermes Framework}
        \label{fig:teaser_a}
    \end{subfigure}
    \begin{subfigure}{\widthof{\includegraphics[height=\myfigheight, trim={365pt 0 251pt 0}, clip]{figures/teaser.pdf}}}
        \centering
        \includegraphics[height=\myfigheight, trim={365pt 0 251pt 0}, clip]{figures/teaser.pdf}
        \caption{Attention Analysis}
        \label{fig:teaser_b}
    \end{subfigure}
    \begin{subfigure}{\widthof{\includegraphics[height=\myfigheight, trim={529pt 0 0 0}, clip]{figures/teaser.pdf}}}
        \centering
        \includegraphics[height=\myfigheight, trim={529pt 0 0 0}, clip]{figures/teaser.pdf}
        \caption{Efficiency Test}
        \label{fig:teaser_c}
    \end{subfigure}
    \caption{\textbf{Left}: \hermes is a training-free approach for efficient streaming video understanding, enabling stable inference by reusing KV cache and performing hierarchical management of video tokens stored in KV cache. \textbf{Middle}: \hermes is based on a mechanistic investigation of the layer-wise attention preferences over hierarchical video information. \textbf{Right}: We evaluate \llava on a single A800 GPU (80 GB). As input frames increase, \hermes consistently maintains extremely low latency (TTFT < 30 ms) and stable GPU memory consumption, exhibiting no risk of OOM errors and requiring no auxiliary external computational resources.}
\end{figure*}

\section{Introduction}
\label{sec:introduction}



Recent years have witnessed remarkable evolution in the capabilities of Multimodal Large Language Models (MLLMs) in video understanding tasks~\cite{gemini25, li2024llavaonevisioneasyvisualtask, bai2025qwen3vltechnicalreport}.
Despite the progress, the rapid emergence of real-time applications demands stable long video understanding, low-latency response, and memory-efficient deployment. 
However, existing MLLMs struggle to simultaneously satisfy these requirements on streaming videos.
Notably, TimeChat-Online~\cite{timechatonline} observes that a large number of streaming video tokens are redundant, motivating compression methods to address these challenges. While numerous compression techniques have been proposed for offline videos~\cite{wang2025videotreeadaptivetreebasedvideo, yang2024visionziplongerbetternecessary, tao2025dycokedynamiccompressiontokens}, most are ill-suited for memory management in streaming scenarios, as streaming inputs are unpredictable in future frames and queries.

To adapt to streaming inputs, recent research introduces specialized memory management techniques, which generally fall into two paradigms: external memory and internal memory. External memory methods store video content as captions or raw vision patches in databases, and perform ad-hoc retrieval and multimodal prefilling at query time~\cite{xiong2025streamingvideounderstandingmultiround, yang2025streamagentanticipatoryagentsstreaming}, suffering from high latency and a lack of end-to-end cohesion. Additionally, many of these methods necessitate costly model-specific training~\cite{wang2025streambridgeturningofflinevideo, xu2025streamingvlmrealtimeunderstandinginfinite, zeng2025streamforestefficientonlinevideo}. In contrast, internalizing memory directly into the key-value cache (KV cache) remains underexplored, yet is crucial for low-latency responses and seamless end-to-end reasoning over stored video contexts. Moreover, KV cache naturally acts as a latent, model-intrinsic memory~\cite{hu2025memoryageaiagents} that frequently interacts with the video stream, making it particularly suitable for training-free memory management. ReKV~\cite{di2025streamingvideoquestionansweringincontext} and LiveVLM~\cite{ning2025livevlmefficientonlinevideo} are representative training-free, cache-based methods for streaming memory management. They store previous video segments in external CPU or disk and need to perform an additional retrieval when a user query arrives, which still rely on external computational resources and leads to significant latency. StreamMem~\cite{yang2025streammemqueryagnostickvcache} leverages chat template tokens to guide compression but lacks fine-grained KV management and mechanistic interpretability.


To overcome the aforementioned limitations of existing streaming video methods, we propose \hermes (KV Cache as \underline{\textbf{H}}i\underline{\textbf{ER}}archical \underline{\textbf{M}}emory for \underline{\textbf{E}}fficient \underline{\textbf{S}}treaming Video Understanding), a training-free and plug-and-play approach that can be seamlessly integrated into existing MLLMs. 
Grounded in a mechanistic investigation of layer-wise attention shown in~\cref{fig:teaser_b}, we conceptualize KV cache as a hierarchical memory framework that stores video information across multiple levels of granularity: 
shallow layers function as sensory memory, exhibiting a strong recency bias toward newly arriving frames; deep layers act as long-term memory, focusing on frame-level rhythmic anchor tokens; and middle layers serve as transitional working memory that balances recency information with frame-level semantic representations.
Our method \hermes comprises three components: \emph{hierarchical KV cache management}, \emph{cross-layer memory smoothing}, and \emph{position re-indexing}. During inference, \hermes reuses the compact KV cache and requires no auxiliary computations or external devices upon the arrival of user queries, thereby guaranteeing real-time responses. Experiments show that \hermes maintains stable and accurate performance with up to 68\% fewer video tokens, while maintaining consistently low response latency and a constant GPU memory footprint.

To summarize, our main contributions are as follows:
\begin{enumerate}[leftmargin=*]
\itemsep0em 
\item
Grounded in a mechanistic analysis on attention visualization, we pioneer the conceptualization of KV cache as a hierarchical video memory framework across multiple granularities.

\item
We propose \hermes, a training-free method for streaming video understanding by reusing hierarchically managed KV cache. Despite reducing video tokens by up to 68\%, \hermes achieves competitive accuracy, with gains of up to 11.4\% on streaming benchmarks.

\item
\hermes exhibits outstanding efficiency in streaming scenarios. Compared to the prior training-free SOTA method, it achieves up to a 10$\times$ speedup in latency. With a constant, compact GPU memory footprint and no auxiliary computation at query time, \hermes ensures consistently low-latency responses.

\end{enumerate}
\begin{figure*}[!t]
  \centering


\begin{subfigure}[t]{0.246\textwidth}
    \centering
    \includegraphics[width=\linewidth]{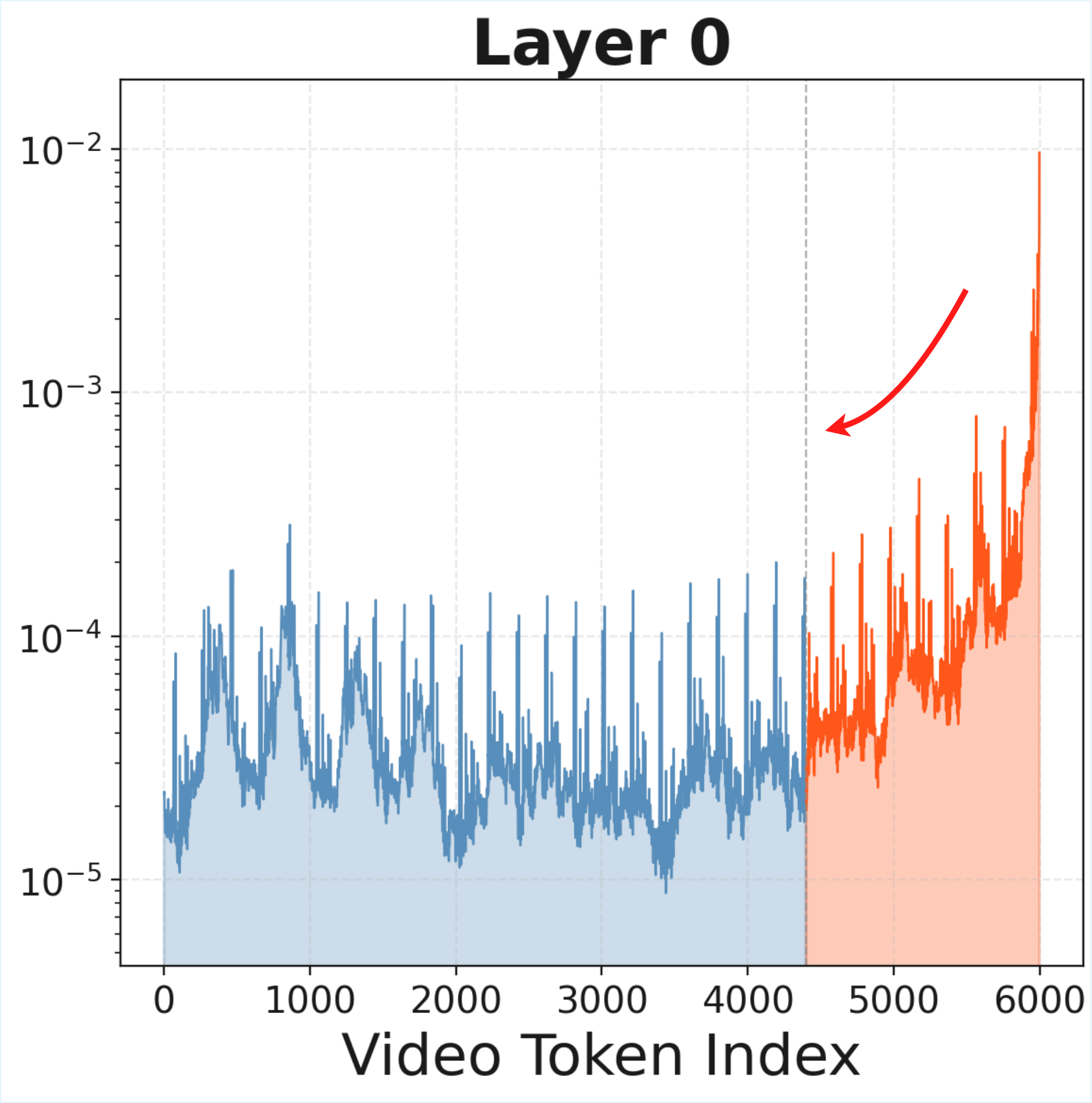}
    \caption{Shallow layer attention.}
    \label{fig:shallow_vis}
  \end{subfigure}
  \begin{subfigure}[t]{0.50\textwidth}
    \centering
    \includegraphics[width=\linewidth]{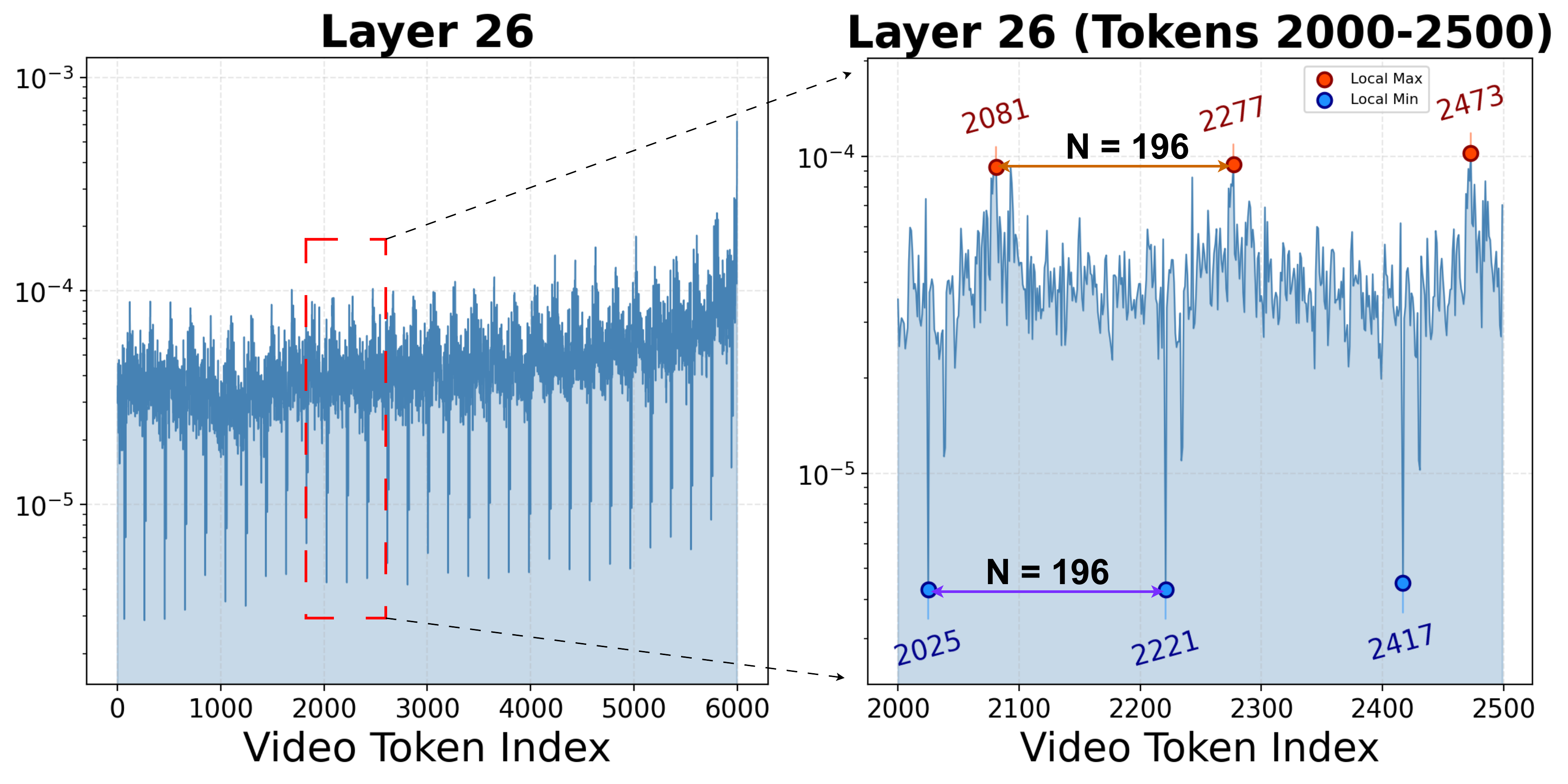}
    \caption{Deep layer attention.}
    \label{fig:deep_vis}
  \end{subfigure}
  \begin{subfigure}[t]{0.242\textwidth}
    \centering
    \includegraphics[width=\linewidth]{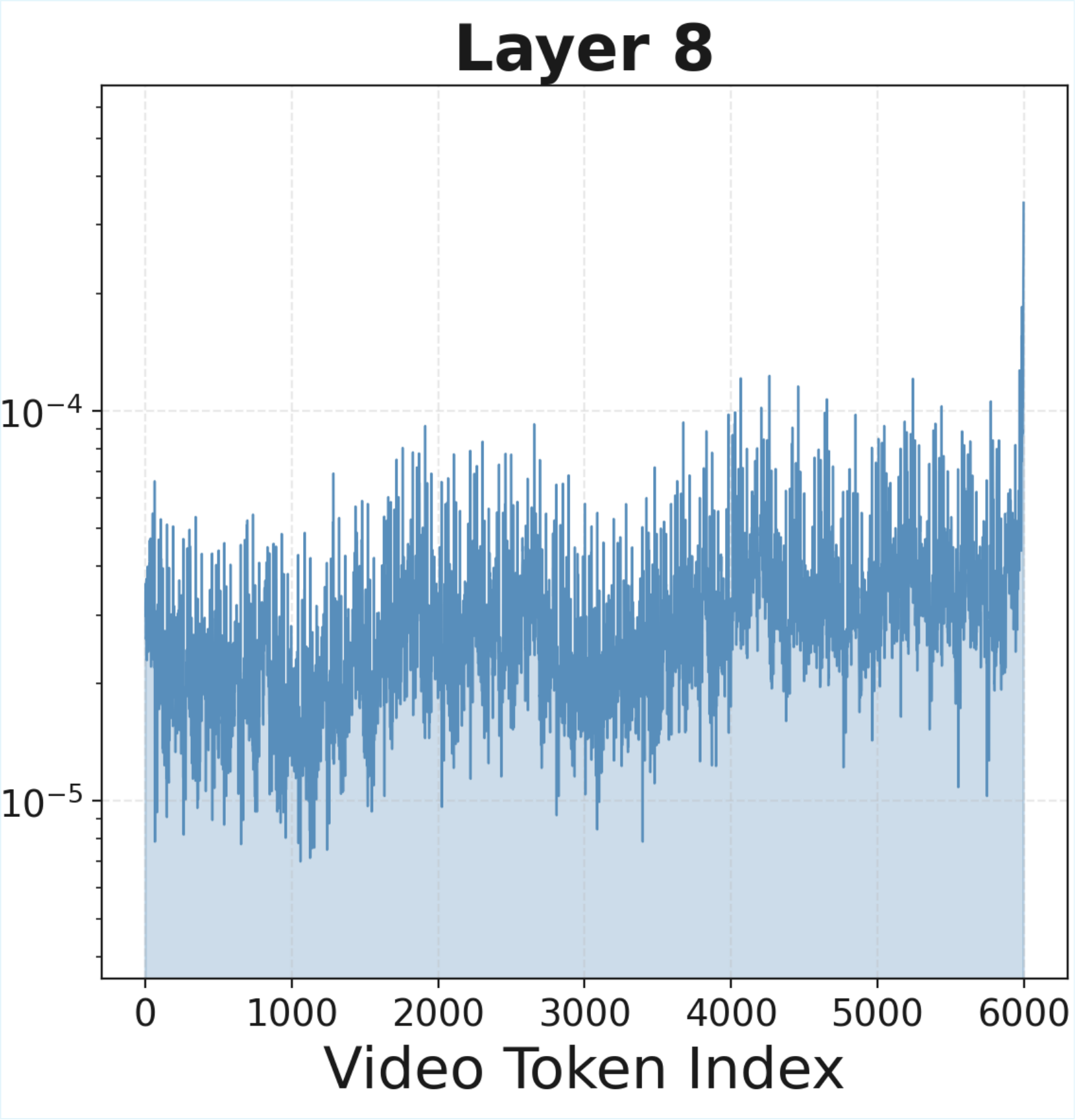}
    \caption{Middle layer attention.}
    \label{fig:mid_vis}
  \end{subfigure}


  \caption{Visualization of the average attention weights (log scale) for user queries over video tokens in \llava with a FIFO KV cache budget of 6K video tokens per layer, averaged across 300 user video questions.}
  \label{fig:layer_vis}
\end{figure*}





\section{Layer-wise Preference for Hierarchical Streaming Video Information}
\label{sec: investigation}


Sliding Window is a standard paradigm for streaming video processing by incrementally encoding the continuous video stream chunk by chunk. When KV cache reaches the pre-defined memory budget, token eviction is triggered, and deciding which tokens to keep is crucial for stable understanding. Existing methods~\cite{di2025streamingvideoquestionansweringincontext, yang2025streammemqueryagnostickvcache, xu2025streamingvlmrealtimeunderstandinginfinite} rely on coarse-grained eviction strategies such as FIFO uniformly across all layers, overlooking layer-wise attention preferences.

To fill this gap, we conduct a mechanistic investigation of attention preferences in MLLM decoder layers, revealing how layers specialize in storing multiple-granularity video memory. To derive generalized insights, we randomly sample 100 video-question pairs from each of the short (62s\footnote{To ensure the sliding window contains 6,000 tokens, a video at 0.5 fps for LLaVA-OV must have a duration of at least $6,000 / 196 / 0.5\approx 62s$.} - 141s), medium (251s - 1,092s) and long (1,795s - 3,579s) duration subsets of the VideoMME benchmark~\cite{fu2025videommefirstevercomprehensiveevaluation} to cover diverse video durations and user queries. The video samples are uniformly sampled at 0.5 fps and subsequently fed into \llava in a streaming chunk-wise manner, with each chunk containing 8 frames. \llava consists of 28 decoder layers, and each video frame is uniformly encoded into 196 visual tokens. During the prefilling stage for video tokens, we maintain a constant budget $|M|$ of 6K video tokens per KV cache layer. After each eviction step, the positional indices of tokens per KV cache layer are re-indexing to contiguous [0, $|M|$).

Layer-wise attention visualizations over video tokens maintained in a FIFO KV cache in~\cref{fig:layer_vis} reveal three general stages of attention preference, along with more visualization results presented in~\cref{app:attn_vis}:

\begin{itemize} [leftmargin=*]
\itemsep0em
\item
\textbf{Shallow Layers as Sensory Memory}:
As shown in~\cref{fig:shallow_vis}, the shallow layers (e.g., layer 0) exhibit an intense recency bias, with attention sharply concentrated on the most recent visual tokens and rapidly decaying over earlier ones. This behavior aligns with the concept of \emph{Sensory Memory}~\cite{ATKINSON196889, shan2025cognitivememorylargelanguage}: shallow layers function as a short-lived buffer for the most recent visual inputs, enabling the model to quickly perceive incoming frames.

\item 
\textbf{Deep Layers as Long-term Memory}: 
In deep layers (e.g., layer 26 in~\cref{fig:deep_vis}), recency bias largely disappears. Instead, the attention pattern becomes highly sparse and rhythmic, with local extrema appearing at regular intervals. These extrema are exactly N = 196 tokens apart, matching to the number of tokens encoding a single frame in \llava. These local maxima can be regarded as frame-level "anchor tokens", summarizing the visual information of each frame. This pattern reflects \emph{Long-term Memory}~\cite{ATKINSON196889, shan2025cognitivememorylargelanguage}: deep layers store critical frame-level semantic representations for long-horizon understanding.

\item 
\textbf{Middle Layers as Working Memory}:
Middle layers (e.g., layer 8 in~\cref{fig:mid_vis}) exhibit a gradual reduction in recency bias, with attention more evenly distributed across recent and earlier tokens. Simultaneously, the attention begins to transition toward the rhythmic patterns in the deep layers. This behavior corresponds to \emph{Working Memory}~\cite{BADDELEY197447, hu2025memoryageaiagents}: middle layers integrate recent and earlier visual information, bridging short-term sensory traces with frame-level semantic summaries.

\end{itemize}

\begin{figure*}[t]
  \centering
    \includegraphics[width=\linewidth]{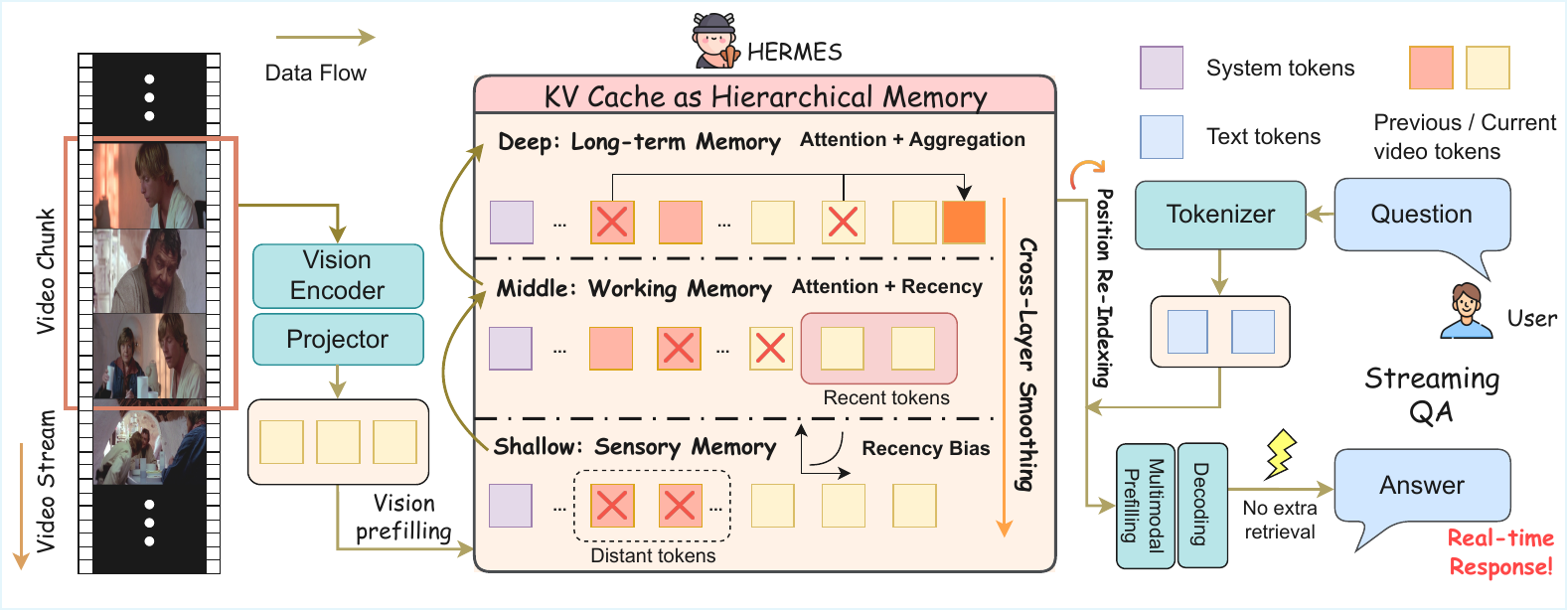}
  \caption{Overview of the \hermes architecture for streaming video QA. By implementing a hierarchical KV cache and specialized management strategies, \hermes enables real-time and accurate responses through direct cache reuse, eliminating the need for additional retrieval operations or external memory whenever users pose questions.}
  \label{fig:hermes}
\end{figure*}

\section{HERMES}
\label{sec:method}
We propose \hermes, a training-free framework that can be seamlessly integrated with MLLMs. As shown in~\cref{fig:hermes}, \hermes has three components: hierarchical KV cache management, cross-layer memory smoothing, and position re-indexing.

\subsection{Hierarchical KV Cache Management}
\label{sec:hierarchical}
Motivated by the layer-wise attention patterns identified in~\cref{sec: investigation}, we design a hierarchical KV cache strategy.
For each video token with KV cache index $i$ at layer $l$, where $i$ denotes its physical position in KV cache, we compute an importance score $S_i^l$ to decide its retention:

\begin{itemize} [leftmargin=*]
\itemsep0em
\item
\textbf{Shallow Layers}:
They act as sensory memory with strong recency bias. Inspired by Ebbinghaus’ memory decay theory~\cite{ebbinghaus2013memory}, we model token importance using an exponential forgetting curve based on temporal distance:
\begin{equation}
\label{eq:shallow}
S_i^l = \alpha_i^l \cdot e^{-k\Delta t_i}, \Delta t_i = T - 1 - i,
\end{equation}
where $T$ is the total number of video tokens in the cache, $k > 0$ is the forgetting rate, $\alpha_i^l$ denotes the normalization factor.

\item
\textbf{Deep Layers}:
Deep layers function as frame-level long-term memory with stable anchor tokens. Their attention distributions are sparse, and these anchor tokens consistently receive high attention across frames, making attention magnitude a reliable indicator of long-term importance. We therefore compute token importance directly from attention weights with respect to the user query. To handle unpredictable queries in streaming scenarios, we use a generic guidance prompt (see~\cref{app:prompt}) as a pseudo query. 
Token importance is computed as:
\begin{equation}
\label{eq:deep}
S_i^l = \alpha_i^l \cdot W_i^l,
\end{equation}
where $W_i^l$ denotes the attention weight of the $i$-th token at the layer $l$.

\item
\textbf{Middle Layers}:
Middle layers serve as working memory, transitioning from recency-dominated shallow layers to attention-driven deep layers. 
We compute importance by interpolating recency and attention with a layer-dependent weight:
\begin{equation}
\omega^l = \omega_0 - \gamma \cdot \frac{l - l_{\text{short}}}{l_{\text{long}} - l_{\text{short}}},
\end{equation}
where $l_{\text{short}}$ and $l_{\text{long}}$ denote the layer indices, with $\omega_0 = 0.75$ and $\gamma = 0.6$.
The importance score of token $i$ at layer $l$ is then computed as
\begin{equation}
S_i^l = (1 - \omega^l)\,A_i^l + \omega^l\, R_i^l,
\end{equation}
where $A_i^l$ and $R_i^l$ denote the normalized attention weight and recency score, respectively, computed as in~\cref{eq:deep,eq:shallow}.

\end{itemize}


\subsection{Cross-Layer Memory Smoothing}
\label{sec: smoothing}
Hierarchical KV cache management may introduce cross-layer inconsistency, as tokens at the same cache index can be evicted independently across layers, leading to misaligned visual memory. Since effective LLM memory relies on cross-layer interaction~\cite{packer2024memgptllmsoperatingsystems, behrouz2024titanslearningmemorizetest, sun2025hierarchicalmemoryhighefficiencylongterm, hu2025memoryageaiagents}, we address this issue with \emph{Cross-Layer Memory Smoothing}.

Instead of treating video tokens at the same KV cache index as independent across layers, we propagate and smooth importance signals from deeper to shallower layers. 
Given raw importance scores $S_i^l$, the smoothed score is computed as:
\begin{equation}
\tilde{S_i^l} = (1 - {\lambda}_l) \cdot S_i^l + \lambda_l \cdot S_i^{l+1},
\end{equation}
$\lambda \in [0,1]$ is the smoothing hyperparameter that controls the strength of cross-layer smoothing.

We then apply Top-K selection based on $\tilde{S}_i^l$ to maintain a fixed memory budget $|M|$ per layer:
\begin{equation}
\begin{aligned}
\mathcal{I}_l &= \mathrm{TopK}(\tilde{S}_l, |M|), \\
K_l &= K_l[\mathcal{I}_l], \quad
V_l = V_l[\mathcal{I}_l].
\end{aligned}
\end{equation}

To preserve long-term information, evicted tokens are aggregated into a \textbf{summary token} per layer, which compactly encodes long-term memory and is retained in the KV cache (see~\cref{alg:summary}).


\subsection{Position Re-Indexing}
\label{sec:pos}
Continuous accumulation of streaming inputs causes positional indices to exceed the model's maximum supported range, severely degrading text generation quality. To stabilize inference, we apply position re-indexing, which remaps positional indices to a contiguous range $[0, |M|)$ within the memory budget $|M|$. We design two strategies:

\paragraph{Lazy Re-Indexing} Re-indexing is triggered only when positional indices approach the model limit, resulting in lower computational overhead. By preserving the original positional indices of recent tokens, it prevents positional drift compared to eager re-indexing, making it well suited for streaming video understanding.

\paragraph{Eager Re-Indexing} Re-indexing is performed at each compression step, maintaining strictly contiguous RoPE indices in KV cache. While this strategy stabilizes long-range visual semantics~\cite{kim2024infinipotinfinitecontextprocessing,kim2025infinipotvmemoryconstrainedkvcache, xu2025streamingvlmrealtimeunderstandinginfinite}, it leads to higher computational cost due to frequent re-indexing, making it more suitable for offline videos.

The details of re-indexing implementation for 1D RoPE (LLaVA-OV) and 3D M-RoPE (Qwen2.5-VL) are illustrated in~\cref{app:1d_pos} and ~\cref{app:3d_pos}, respectively.

\section{Experiments}
\label{sec:experiments}

\subsection{Experimental Setup}
\paragraph{Benchmarks.}
We evaluate \hermes on diverse streaming and offline benchmarks. For streaming understanding, we use StreamingBench~\cite{lin2024streamingbenchassessinggapmllms}, OVO-Bench~\cite{li2025ovobenchfarvideollmsrealworld} and RVS (including RVS-Ego and EVS-Movie)~\cite{zhang2024flashvstreammemorybasedrealtimeunderstanding}.
For offline video evaluation, we adopt one short video dataset MVBench~\cite{li2024mvbenchcomprehensivemultimodalvideo}, along with two long video datasets, VideoMME~\cite{fu2025videommefirstevercomprehensiveevaluation} and Egoschema~\cite{mangalam2023egoschemadiagnosticbenchmarklongform}. We conduct evaluation on the official dev split of Egoschema and report VideoMME results without subtitles. Our benchmark selection covers both multiple-choice and open-ended questions as QA form. The details of utilized benchmarks are demonstrated in~\cref{app:details_of_benchmarks}.

\paragraph{Models.}
To further verify the broad applicability of our method, we select two popular open-source MLLM series, LLaVA-OneVision (LLaVA-OV)~\cite{li2024llavaonevisioneasyvisualtask} and Qwen2.5-VL~\cite{bai2025qwen25vltechnicalreport}. Each is tested across two different parameter scales, covering a large range from 0.5B to 32B. For Qwen2.5-VL, we maintain its native dynamic resolution on video input, ensuring a fair comparison with the base model.

\paragraph{Implementation Details.}
For evaluating \hermes across all benchmarks, each video is encoded and processed chunk by chunk, with 16 frames per chunk, and sequentially prefilling the backbone LLM. Then, token compression is triggered once the predefined memory budget is exceeded.

For the layer partition, we follow the mechanistic investigations presented in ~\cref{sec: investigation}: 10\% shallow, 60\% middle and 30\% deep layers. A more comprehensive analysis of attention behaviors as supportive evidence can be found in~\cref{fig:more_vis}. The cross-layer memory smoothing hyperparameter $\lambda$ proposed in~\cref{sec: smoothing} is layer-dependent, with detailed configurations reported in~\cref{app:smooth_config}.

All evaluations are conducted using FP16 mixed precision and efficiency tests are conducted on a single A800 GPU, consistent with prior works~\cite{di2025streamingvideoquestionansweringincontext, chen2025streamingtomstreamingtokencompression}. Greedy decoding is used to generate deterministic outputs. Accuracy evaluations can be completed on one H200 GPU.

\subsection{Main Results}
\paragraph{Streaming Video Understanding}
\begin{table*}[h!]
\renewcommand{\arraystretch}{0.9}
    \centering
    \footnotesize
    \caption{Performance comparison (\%) on StreamingBench and OVO-Bench. The "Avg." column reports the results of the average accuracy of real-time visual perception and backward tracing tasks.}
    \label{tab:streaming_main}
    \begin{tabular}{l c | c c c c}
    \toprule
    \multirow{2}{*}{\textbf{Model}} & \multirow{2}{*}{\textbf{\#Frames}} & \textbf{StreamingBench} & \multicolumn{3}{c}{\textbf{OVO-Bench}} \\
    & & Real-Time & Real-Time & Backward & \textbf{Avg.} \\
    \midrule
    \midrule
    Human & - & 91.46 & 93.20 & 92.33 & 92.83 \\
    \midrule
    \multicolumn{6}{c}{\textbf{Proprietary MLLMs}} \\
    \midrule
    Gemini 1.5 pro~\cite{gemini25} & 1 fps & 75.69 & 69.32 & 62.54 & 66.41\\
    GPT-4o~\cite{openai2024gpt4ocard} & 64 & 73.28 & 64.46 & 60.75 & 62.87\\
    Claude 3.5 Sonnet~\cite{claude3_5} & 20 & 72.44 & - & - & -\\
    \midrule
    \multicolumn{6}{c}{\textbf{Open-source Offline MLLMs}} \\
    \midrule
    Video-LLaMA2-7B~\cite{videollama2} & 32 & 49.52 & - & - & -\\
    VILA-1.5-8B~\cite{vila} & 14 & 52.32 & - & - & -\\
    Video-CCAM-14B~\cite{videoccam} & 96 & 53.96 & - & - & -\\
    LongVA-7B~\cite{longva} & 128 & 59.96 & - & - & -\\
    Qwen2-VL-7B~\cite{wang2024qwen2vlenhancingvisionlanguagemodels} & 64 & 69.04 & 60.65 & 48.58 & 54.62 \\
    InternVL-V2-8B~\cite{internvl2} & 16 & 63.72 & 60.73 & 44.00 & 52.37 \\
    LLaVA-NeXT-Video-32B~\cite{llava-next} & 64 & 66.96 & - & - & -\\
    MiniCPM-V-2.6-8B~\cite{minicpm} & 32 & 67.44 & - & - & -\\
    \addlinespace[1pt]
    
    \midrule
    \multicolumn{6}{c}{\textbf{Open-source  Online MLLMs}} \\
    \midrule
    Flash-VStream-7B~\cite{flashvstream} & 1 fps & 23.23 & 29.86 & 25.35 & 27.61\\
    VideoLLM-online-8B~\cite{videollmonline} & 2 fps & 35.99 & 20.79 & 17.73 & 19.26\\
    Dispider-7B~\cite{dispider} & 1 fps & 67.63 & 54.55 & 36.06 & 45.31\\
    TimeChat-Online-7B~\cite{timechatonline} & 1 fps & 75.36 & 61.90 & 41.70 & 51.80\\
    StreamForest-7B~\cite{zeng2025streamforestefficientonlinevideo} & 1 fps & 77.26 & 61.20 & 52.02 & 56.61 \\

    \midrule
    \multicolumn{6}{c}{\textbf{Training-free Offline-to-Online Methods}} \\
    \midrule
    LLaVA-OV-7B~\cite{li2024llavaonevisioneasyvisualtask} & 64 & 71.34 & 63.06 & 43.64 & 53.35 \\

    \hspace{3pt} + ReKV~\cite{di2025streamingvideoquestionansweringincontext} & 0.5 fps &  69.22 & 57.33 & 44.16 & 50.75 \\

    \hspace{3pt} + LiveVLM~\cite{ning2025livevlmefficientonlinevideo} & 0.5 fps & 72.92 & - & - & -\\

    \hspace{3pt} + StreamKV~\cite{chen2025streamkvstreamingvideoquestionanswering} & 0.5 fps & 68.80 & - & - & -\\

    \rowcolor{gray!20}  \hspace{3pt} + HERMES (6K tokens) & 0.5 fps & 72.63 & 65.07 & 48.80 & 56.94 \\

    \rowcolor{gray!40}  \hspace{3pt} + HERMES (4K tokens) & 0.5 fps & \textbf{73.23} & \textbf{66.34} & \textbf{50.20} & \textbf{58.27} \\
    
    \midrule

    LLaVA-OV-0.5B~\cite{li2024llavaonevisioneasyvisualtask} & 64 & 59.64 & 49.70 & 34.59 & 42.15 \\

    \hspace{3pt} + ReKV~\cite{di2025streamingvideoquestionansweringincontext} & 0.5 fps &  57.39 & 43.77 & 33.06 & 38.42 \\

    \rowcolor{gray!20} \hspace{3pt} + HERMES (6K tokens) & 0.5 fps & 61.04 & 50.34 & 34.75 & 42.55 \\

     \rowcolor{gray!40} \hspace{3pt} + HERMES (4K tokens) & 0.5 fps & \textbf{62.04} & \textbf{50.72} & \textbf{34.80} & \textbf{42.76} \\
    
    \midrule    
    Qwen2.5-VL-7B~\cite{bai2025qwen25vltechnicalreport} & 1 fps & 73.31 & 59.90 & 44.65 & 52.28 \\
    \rowcolor{gray!20} \hspace{3pt} + HERMES (6K tokens) & 1 fps & 78.72 & 68.42 & 48.10 & 58.26 \\
    \rowcolor{gray!40}\hspace{3pt} + HERMES (4K tokens) & 1 fps & \textbf{79.44} & \textbf{68.98} & \textbf{49.43} & \textbf{59.21} \\
    
    \midrule
    
    Qwen2.5-VL-32B~\cite{bai2025qwen25vltechnicalreport} & 1 fps & 74.27 & 64.40 & 50.33 & 57.37 \\
    \rowcolor{gray!20}\hspace{3pt} + HERMES (6K tokens) & 1 fps & \textbf{80.20} & 71.93 & \textbf{57.71} & \textbf{64.82} \\
    \rowcolor{gray!40}\hspace{3pt} + HERMES (4K tokens) & 1 fps & 80.08 & \textbf{72.37} & 55.42 & 63.90 \\
    
    \midrule
    
    Qwen3-VL-8B~\cite{bai2025qwen3vltechnicalreport} & 2 fps & 78.92 & 68.64 & 47.03 & 57.84 \\
    \rowcolor{gray!20}\hspace{3pt} + HERMES (6K tokens) & 2 fps & \textbf{81.32} & 73.21 & 46.78 & 60.00 \\
    \rowcolor{gray!40}\hspace{3pt} + HERMES (4K tokens) & 2 fps & 81.28 & \textbf{73.29} & \textbf{49.28} & \textbf{61.29} \\

    \midrule
    
    Qwen3-VL-4B~\cite{bai2025qwen3vltechnicalreport} & 2 fps & 78.32 & 70.67 & 50.05 & 60.36 \\
    \rowcolor{gray!20}\hspace{3pt} + HERMES (6K tokens) & 2 fps & \textbf{78.40} & 71.90 & 54.00 & 62.95 \\
    \rowcolor{gray!40}\hspace{3pt} + HERMES (4K tokens) & 2 fps & 78.24 & \textbf{72.32} & \textbf{55.03} & \textbf{63.68} \\
    
    \bottomrule
    
    \end{tabular}
\end{table*}

\begin{table}[t]
    \centering
    \begin{minipage}[t]{0.49\textwidth}
      \centering
      \small
      \caption{Performance on RVS-Ego and RVS-Movie. \dag: ReKV caches the KV states of all previously seen frames and is therefore treated as an upper bound.}
      \label{tab:rvs}
      \resizebox{\linewidth}{!}{
      \begin{tabular}{lcccc}
        \toprule
        \multirow{2}{*}{\textbf{Model}} & \multicolumn{2}{c}{\textbf{RVS-Ego}} & \multicolumn{2}{c}{\textbf{RVS-Movie}} \\
        \cmidrule(lr){2-3} \cmidrule(lr){4-5}
         & \textbf{Acc} & \textbf{Score} & \textbf{Acc} & \textbf{Score} \\
        \midrule
        \llava~\cite{li2024llavaonevisioneasyvisualtask} & 56.2 & 3.7 & 43.0 & 3.3 \\
        \hspace{3pt}+ ReKV$^{\dag}$ \citep{di2025streamingvideoquestionansweringincontext} & 63.7 & 4.0 & 54.4 & 3.6  \\
        \hspace{3pt}+ ReKV w/o off.~\cite{di2025streamingvideoquestionansweringincontext} & 55.8 & 3.3 & 50.8 & 3.4 \\
        \hspace{3pt}+ Flash-VStream \citep{flashvstream} & 57.0 & 4.0 & 53.1 & 3.3 \\
        \hspace{3pt}+ InfiniPot-V \citep{kim2025infinipotvmemoryconstrainedkvcache} & 57.9 & 3.5 & 51.4 & 3.5 \\
        \hspace{3pt}+ StreamMem \citep{yang2025streammemqueryagnostickvcache} & 57.6 & 3.8 & 52.7 & 3.4 \\
        \hspace{3pt}+ StreamingTOM~\cite{chen2025streamingtomstreamingtokencompression} & 58.3 & 3.9 & 53.2 & 3.5 \\
        \rowcolor{gray!20}\hspace{3pt}+ HERMES (6K tokens) & \textbf{60.3} & \textbf{4.0} & \textbf{54.4} & \textbf{3.6} \\
        \rowcolor{gray!40}\hspace{3pt}+ HERMES (4K tokens) & 58.3 & 3.9 & \textbf{54.4} & \textbf{3.6} \\
        \midrule
        
        \llavasmall~\cite{li2024llavaonevisioneasyvisualtask} & 51.8 & 3.7 & 37.2 & 3.2 \\
        \hspace{3pt}+ ReKV$^{\dag}$ \citep{di2025streamingvideoquestionansweringincontext} & 54.7 & 3.9 & 44.6 & 3.4  \\
        \rowcolor{gray!20}\hspace{3pt}+ HERMES (6K tokens) & \textbf{53.0} & \textbf{3.8} & \textbf{42.5} & \textbf{3.4} \\
        \rowcolor{gray!40}\hspace{3pt}+ HERMES (4K tokens) & 52.7 & \textbf{3.8} & 41.7 & \textbf{3.4} \\
        \bottomrule
        
      \end{tabular}
      }
    \end{minipage}%
    \hfill%
    \begin{minipage}[t]{0.49\textwidth}
        \centering
        \small
        \caption{Efficiency across input frame numbers under two chunk sizes. "TTFT" denotes \textit{Time to First Token} and "TPOT" denotes \textit{Time Per Output Token}.}
        \label{tab:efficiency_chunk}
        \resizebox{\linewidth}{!}{
        \begin{tabular}{l cccc}
        \toprule
        \multirow{2}{*}{\textbf{Metric}} & \multicolumn{4}{c}{\textbf{Frames}} \\
         & \textbf{16} & \textbf{64} & \textbf{256} & \textbf{512}\\
        \midrule
        \multicolumn{5}{c}{\textit{\textbf{Chunk Size: 8}}} \\
        GPU Mem. / GB $\downarrow$          & 16.54 & 16.66 & 16.66 & 16.66  \\
        TTFT / ms $\downarrow$               & 27.01 & 28.41 & 28.44 & 28.41  \\
        TPOT / ms $\downarrow$  & 24.43 & 23.89 & 24.02 & 23.98 \\
        \midrule
        \multicolumn{5}{c}{\textit{\textbf{Chunk Size: 16}}} \\
        GPU Mem. / GB $\downarrow$          & 17.46 & 17.66 & 17.66 & 17.66 \\
        TTFT / ms $\downarrow$               & 27.02 & 28.97 & 28.50 & 28.38 \\
        TPOT / ms $\downarrow$  & 24.50 & 23.59 & 23.56 & 23.63 \\
        \bottomrule
        \end{tabular}
        }
    \end{minipage}
\end{table}

Extensive experiments on streaming benchmarks reveal the key findings: 

\noindent
(1) \textit{\hermes outperforms on multiple-choice streaming datasets, showing exceptional real-time understanding and backward tracing capabilities}.
As shown in~\cref{tab:streaming_main}, it achieves state-of-the-art performance on StreamingBench and OVO-Bench, significantly surpassing base models and training-free baselines. Built on \qwen, \hermes reaches 79.44\% and 59.21\% accuracy using only 4K video tokens, improving over \qwen by 6.13\% and 6.93\%, while outperforming all 7B-scale open-source online and offline models. Full results on StreamingBench and OVO-Bench are shown in~\cref{tab:streamingbench_full} and~\cref{tab:ovobench_full} respectively.

\noindent
(2) \textit{\hermes excels on open-ended streaming tasks, showing fine-grained temporal and spatial comprehension}. On RVS-Ego and RVS-Movie (\cref{tab:rvs}), we evaluate the model answer by GPT-3.5-turbo-0125 on accuracy and score (1–5 scale), consistent with compared baselines.
\hermes consistently surpasses all prior training-free methods and improves accuracy by up to 11.4\% over the base model with uniformly sampled 64 frames. These extensive experiments demonstrate \hermes's strong abilities in various streaming tasks, as well as its general applicability across foundation models. Moreover, we provide case studies from RVS benchmark, showing finer-grained temporal (shown in~\cref{fig:case_temporal}) and spatial understanding (shown in ~\cref{fig:case_spatial}) abilities of \hermes than its base model.

\paragraph{Offline Video Understanding}
\begin{table}[t]
    \centering
    \caption{Performance comparison (\%) on offline benchmarks.}
    \begin{tabular}{l c | c c c c}
    \toprule
    \multirow{2}{*}{\textbf{Model}} & \multirow{2}{*}{\textbf{\#Frames}} & \textbf{MVBench} & \textbf{Egoschema} & \multicolumn{2}{c}{\textbf{VideoMME}} \\
    & & & & Long & \textbf{Avg.} \\
    \midrule
    \midrule

    \multicolumn{6}{c}{\textbf{Proprietary MLLMs}} \\
    \midrule
    Gemini 1.5 pro~\cite{gemini25} & 1 fps & 75.69 & 69.32 & 62.54 & 66.41\\
    GPT-4o~\cite{openai2024gpt4ocard} & 64 & 73.28 & 64.46 & 60.75 & 62.87\\
    Claude 3.5 Sonnet~\cite{claude3_5} & 20 & 72.44 & - & - & -\\
    \midrule
    \multicolumn{6}{c}{\textbf{Open-source Offline MLLMs}} \\
    \midrule
    Video-LLaMA2-7B~\cite{videollama2} & 32 & 49.52 & - & - & -\\
    VILA-1.5-8B~\cite{vila} & 14 & 52.32 & - & - & -\\
    Video-CCAM-14B~\cite{videoccam} & 96 & 53.96 & - & - & -\\
    LongVA-7B~\cite{longva} & 128 & 59.96 & - & - & -\\
    LLaVA-Video-7B~\cite{zhang2025llavavideovideoinstructiontuning} & 32 & 58.60 & 57.30 & - & 63.30 \\
    Qwen2-VL-7B~\cite{wang2024qwen2vlenhancingvisionlanguagemodels} & 64 & 67.00 & 66.70 & - & 63.30 \\
    InternVL-V2-8B~\cite{internvl2} & 16 & 65.80 & - & - & 56.30 \\
    Kangaroo-7B~\cite{kangaroo} & 64 & 64.60 & - & - & -\\
    LLaVA-NeXT-Video-32B~\cite{llava-next} & 64 & 66.96 & - & - & -\\
    MiniCPM-V-2.6-8B~\cite{minicpm} & 32 & 67.44 & - & - & -\\

    \addlinespace[1pt]
    
    \midrule
    \multicolumn{6}{c}{\textbf{Open-source  Online MLLMs}} \\
    \midrule
    Dispider-7B~\cite{dispider} & 1 fps & - & 55.60 & - & 57.20\\
    TimeChat-Online-7B~\cite{timechatonline} & 1 fps & 75.36 & 61.90 & 41.70 & 53.22\\
    StreamForest-7B~\cite{zeng2025streamforestefficientonlinevideo} & 1 fps & 70.20 & - & - & 61.40\\
    \midrule
    \multicolumn{6}{c}{\textbf{Training-free Offline-to-Online Methods}} \\
    \midrule
    LLaVA-OV-7B~\cite{li2024llavaonevisioneasyvisualtask} & 64 & \textbf{57.02} & 59.93 & 48.00 & 57.67 \\

    \hspace{3pt} + ReKV~\cite{di2025streamingvideoquestionansweringincontext} & 0.5 fps &  56.83 & \textbf{60.70} & 46.89 & 57.74 \\

    \rowcolor{gray!20}  \hspace{3pt} + HERMES (6K tokens) & 0.5 fps & 56.95 & 60.23 & 49.11 & 58.44 \\

     \rowcolor{gray!40}  \hspace{3pt} + HERMES (4K tokens) & 0.5 fps & 56.92 & 60.29 & \textbf{49.22} & \textbf{58.85} \\
    
    \midrule    
    Qwen2.5-VL-7B~\cite{bai2025qwen25vltechnicalreport} & 1 fps & 65.00 & 58.47 & 53.89 & \textbf{64.52}\\
    \rowcolor{gray!20}\hspace{3pt} + HERMES (6K tokens) & 1 fps & 65.40 & 59.47 & \textbf{54.44} & 62.00 \\
    \rowcolor{gray!40} \hspace{3pt} + HERMES (4K tokens) & 1 fps & \textbf{65.53} & \textbf{59.97} & 53.44 & 60.63 \\

    \bottomrule
    \end{tabular}
    \label{tab:offline_main}
\end{table}

The results presented in~\cref{tab:offline_main} demonstrate the \textit{competitive performance of \hermes across multiple temporal scales on offline benchmarks}, compared to the base model and other training-free methods. Under a limited budget of video tokens, \hermes achieves performance that is better than or comparable to the corresponding base models. \hermes based on \llava surpasses the base model on long video datasets Egoschema and VideoMME, achieving 60.29\% and 58.85\%, respectively, and attains 56.92\% accuracy on the short video dataset MVBench, which is comparable to the base model's 57.02\%.

\subsection{Efficiency Analysis}
\begin{figure}[ht]
  \centering
    \includegraphics[width=\linewidth]{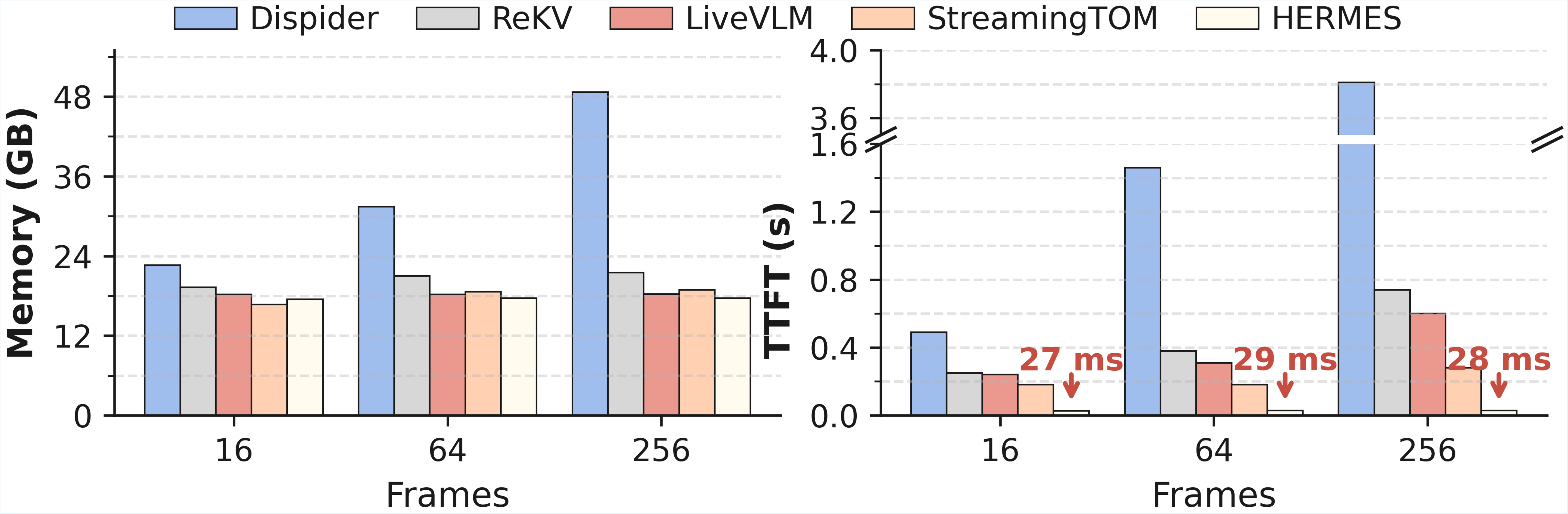}
  \caption{GPU memory and TTFT latency comparison across input frame numbers. \hermes achieves 10 $\times$ faster in TTFT compared to prior SOTA.}
  \label{fig:efficency_compare}
\end{figure}

To evaluate the efficiency of \hermes, we utilize three metrics: peak GPU memory usage, Time to First Token (TTFT), defined as the latency measured from the moment a user inputs a query to the decoding of the first output token, and Time Per Output Token (TPOT) across varying numbers of input frames. All experiments are conducted using \llava as the base model with a 4K-token memory budget. ~\cref{fig:efficency_compare} shows the comparison of memory usage and TTFT among \hermes and representative streaming methods. Unlike Dispider and LiveVLM, \hermes consistently maintains stable memory usage and TTFT as frames increase. Notably, under the 256-frame setting, \hermes achieves 1.04$\times$ reduction in peak memory compared to the prior SOTA LiveVLM, while achieving an impressive 10$\times$ speedup in TTFT over the prior SOTA StreamingTOM.

We further examine the efficiency of \hermes under varying encoded video chunk sizes, with the results shown in~\cref{tab:efficiency_chunk}. GPU memory usage does not increase with longer video lengths due to the fixed memory budget. TTFT and TPOT remain consistently low across varying video lengths and encoding chunk sizes, confirming real-time responsiveness in practical streaming scenarios.

\subsection{Ablation Study}

We conduct ablation studies to evaluate the contributions of \hermes's components and hyperparameter choices, covering: (1) total memory budget, (2) layer-dependent memory budget (3) cross-layer memory smoothing and its hyperparameters, (4) position re-indexing strategies for streaming and offline datasets, (5) guidance prompts and (6) summary tokens for long-term memory retention.

\paragraph{Total Memory Budget}
\label{sec:memory_ablation}
\begin{figure}[ht]
\begin{subfigure}[ht]{0.47\linewidth}
    \centering
    \includegraphics[width=\linewidth]{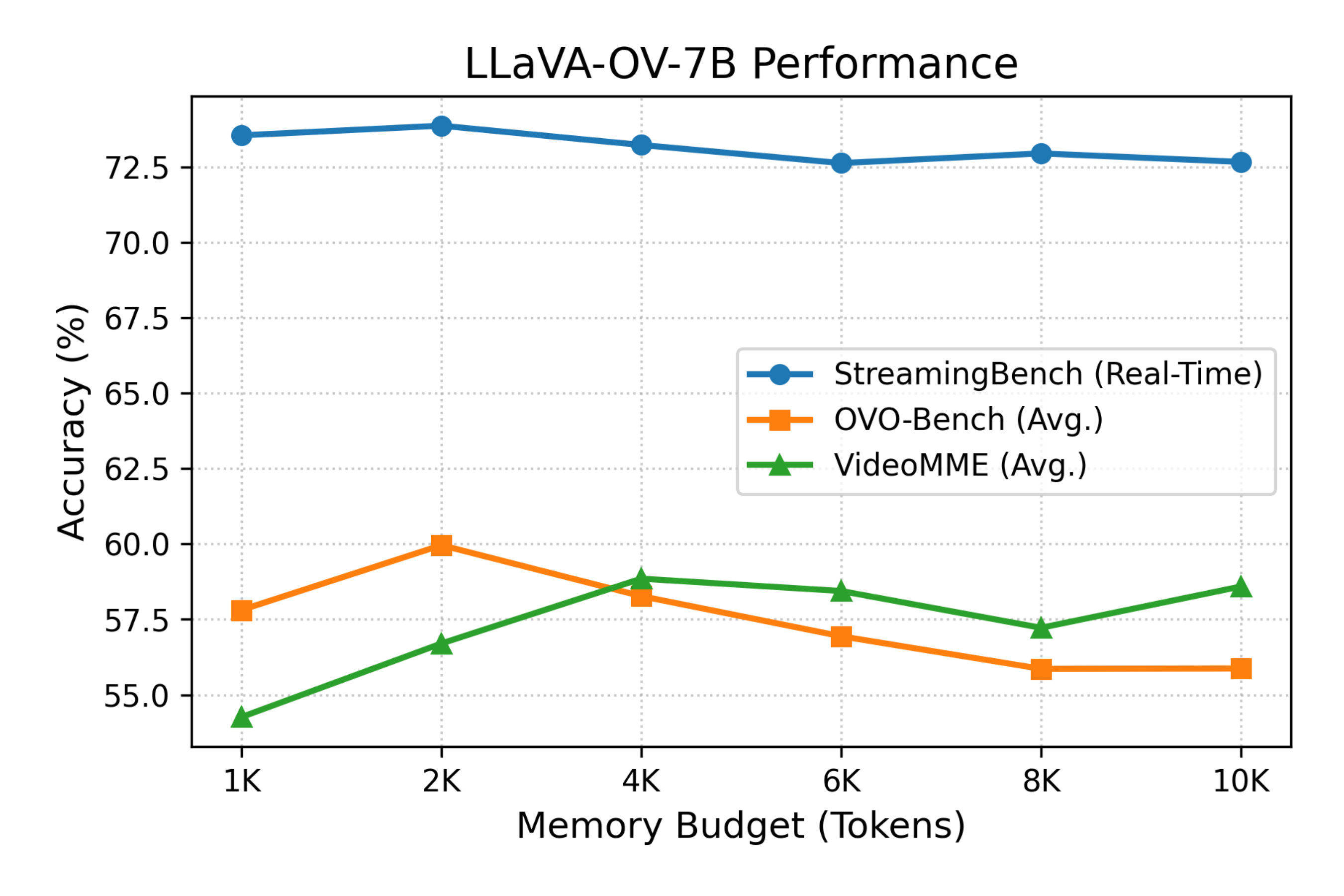}
    \caption{Performance comparison of \llava across different memory budgets.}
    \label{fig:memory_budget_llava}
  \end{subfigure}
  \quad
  \begin{subfigure}[ht]{0.47\linewidth}
    \centering
    \includegraphics[width=\linewidth]{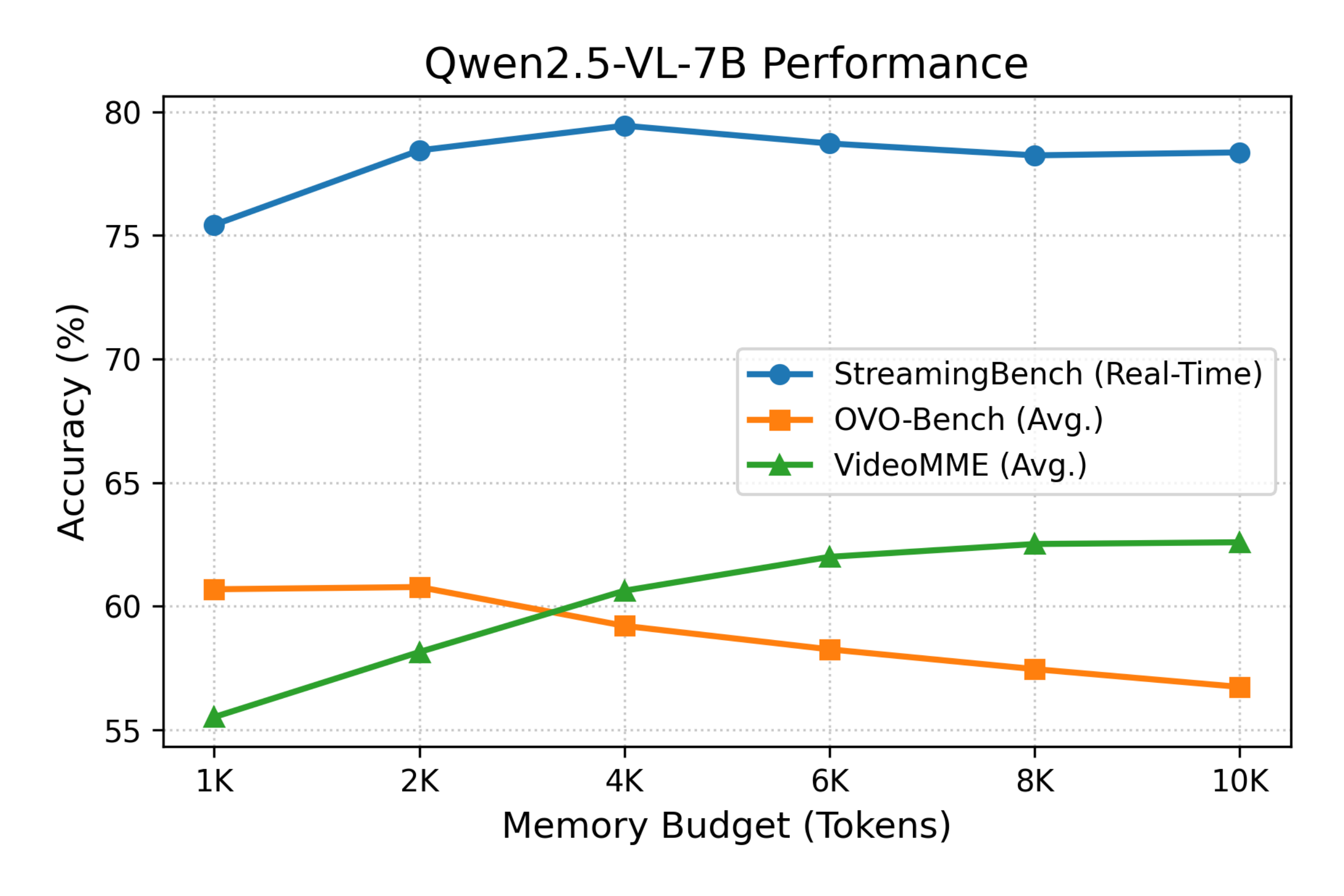}
    \caption{Performance Comparison of \qwen across Different Memory Budgets.}
    \label{fig:memory_budget_qwen}
  \end{subfigure}
\end{figure}

To investigate the impact of total memory budget on understanding performance, we conduct ablations by varying the memory budget $|M|$ from 1K to 10K. As shown in~\cref{fig:memory_budget_llava}, for HERMES built upon LLaVA-OV-7B, the performance on both streaming and offline datasets stabilizes once memory budget reaches 4K. Notably, streaming datasets can tolerate a smaller memory budget. In contrast, the performance on long offline datasets degrades significantly when the memory budget is below 4K. The additional ablation on \qwen is provided in~\cref{fig:memory_budget_qwen}, yielding conclusions consistent with those on \llava.

\paragraph{Layer-dependent Memory Budget}
\label{sec:layer_memory_ablation}
\begin{table*}[htbp]
  \centering
  \small
  \caption{Ablation on layer-dependent budgets.}
  \label{tab:layer_memory_ablation}
  \setlength{\tabcolsep}{3pt}
  \begin{tabular}{ccc c ccc cccc}
    \toprule
    \multicolumn{3}{c}{\textbf{Budget Weight}} & \textbf{StreamingBench} & \multicolumn{3}{c}{\textbf{OVO-Bench}} & \multicolumn{4}{c}{\textbf{VideoMME}} \\
    \cmidrule(lr){1-3} \cmidrule(lr){4-4} \cmidrule(lr){5-7} \cmidrule(lr){8-11}
    Shallow & Middle & Deep & Real-Time & Real-Time & Backward & \textbf{Avg.} & Short & Medium & Long & \textbf{Avg.} \\
    \midrule
    1.0 & 1.0 & 1.0 & 72.63 & 65.07 & 48.80 & 56.94 & 71.33 & 54.89 & 49.11 & 58.44 \\
    1.3 & 1.0 & 0.7 & 72.42 & 65.28 & \textbf{49.27} & \textbf{57.28} & 70.78 & 53.11 & 48.89 & 57.59 \\
    0.7 & 1.6 & 0.7 & \textbf{72.95} & 64.63 & 48.17 & 56.40 & 70.67 & 54.11 & 47.67 & 57.48 \\
    0.7 & 1.0 & 1.3 & 72.79 & \textbf{65.38} & 48.62 & 57.00 & \textbf{71.44} & \textbf{56.00} & \textbf{49.78} & \textbf{59.07} \\
    \bottomrule
  \end{tabular}
\end{table*}
We conduct an ablation study on layer-dependent budgets, where the total token budget remains fixed. The allocation strategy for layer-dependent budgets is described as follows:
Given a fixed total memory budget $|M| = 6000 \times L$, we allocate per-layer budgets proportionally to normalized weights:
$m_i = \left\lfloor \frac{w_i}{\sum_j w_j} \, |M| \right\rfloor$.
The rounding residue $|M| - \sum_i m_i$ is added to the last layer to ensure $\sum_i m_i = |M|$, where $w_i$ is the budget weight of the $i$-th layer, $L$ is the number of layers. The results in~\cref{tab:layer_memory_ablation} show comparable overall performance across configurations, indicating that HERMES is not highly sensitive to the exact layer-dependent budget allocation strategy. Notably, allocating more tokens to deep layers leads to better preservation of long-term memory and improved performance on the long video subset of VideoMME, which is consistent with our observation on layer-wise attention.

\begin{table}[t]
    \centering
    \begin{minipage}[t]{0.47\textwidth}
    \centering
    \small
    \caption{Ablation on different cross-layer memory smoothing hyperparameter $\lambda$.}
    \resizebox{\linewidth}{!}{
    \begin{tabular}{ccc | cccc}
        \toprule
         \multicolumn{3}{c|}{\textbf{Hyperparameter}} & \multicolumn{4}{c}{\textbf{VideoMME}} \\
        $\lambda_{deep}$ & $\lambda_{mid}$ & $\lambda_{shallow}$ & Short & Medium & Long & \textbf{Avg.}\\
        \midrule
        0 & 0 & 0   & 69.67 & 51.11 & 43.44 & 54.74 \\
        0.5 & 0 & 0   & 69.67 & 51.44 & 43.56 & 54.89 \\
        0 & 0.5 & 0   & 70.89 & 54.78 & 46.44 & 57.37 \\
        0 & 0 & 0.5   & 70.89 & 54.44 & 47.00 & 57.44 \\
        0.5 & 0.5 & 0.5   & \textbf{71.78} & 54.78 & 47.33 & 57.96 \\
        \rowcolor{gray!40}0.4 & 0.3 & 0.1   & 71.33 & \textbf{54.89} & \textbf{49.11} & \textbf{58.44} \\
        \bottomrule
    \end{tabular}
    }
    \label{tab:lambda_ablation}
    \end{minipage}
    \quad
    \begin{minipage}[t]{0.47\textwidth}
        \centering
        \small
        \caption{Ablation on summary tokens in deep layers. The gray row is our default setting in all experiments.}
        \resizebox{\linewidth}{!}{
        \begin{tabular}{lc|cccc}
        \toprule
         \multirow{2}{*}{\textbf{Model}} & \multirow{2}{*}{\textbf{Aggregation}} & \multicolumn{4}{c}{\textbf{VideoMME}} \\
        & & Short & Medium & Long & \textbf{Avg.}\\
        \midrule
        \llava & - & 69.89 & 55.11 & 48.00 & 57.67 \\
        \hspace{3pt}+ HERMES & w/o & 71.33 & 54.78 & 47.78 & 57.96 \\
        \rowcolor{gray!40} \hspace{3pt}+ HERMES & w/ & \textbf{71.33} & 54.89 & \textbf{49.11} & \textbf{58.44} \\
        \bottomrule
    \end{tabular}
    }
    
    \label{tab:summary_token_ablation}
    \end{minipage}
\end{table}
\paragraph{Cross-Layer Memory Smoothing}
In~\cref{tab:lambda_ablation}, we evaluate variants without the proposed cross-layer memory smoothing mechanism, as well as alternative hyperparameter configurations. All these variants exhibit degraded performance on the VideoMME benchmark, demonstrating both the critical role of memory smoothing and the effectiveness of our chosen hyperparameter settings.

\begin{table}[ht]
    \centering
    \begin{minipage}[t]{0.47\textwidth}
        \centering
        \small
         \caption{Ablation on different re-indexing strategies on streaming benchmarks. The gray row represents our default setting in all evaluations for streaming benchmarks. "StrBench" represents \textit{StreamingBench}.}
        \label{tab:pos_ablation_streaming}
        \resizebox{\linewidth}{!}{
        \begin{tabular}{lc|c|ccc}
        \toprule
         \multirow{2}{*}{\textbf{Model}} & \multirow{2}{*}{\textbf{Re-Indexing}} & \textbf{StrBench} & \multicolumn{3}{c}{\textbf{OVO-Bench}} \\
        & & Real-Time & Real-Time & Backward & \textbf{Avg.}\\
        \midrule
        \llava & - & 71.34 & 63.06 & 43.64 & 53.35 \\
        \rowcolor{gray!40} \hspace{3pt}+ HERMES & lazy & \textbf{72.63} & \textbf{65.07} & \textbf{48.80} & \textbf{56.94} \\
        \hspace{3pt}+ HERMES & eager & 72.30 & 64.91 & 47.21 & 56.06 \\
        \bottomrule
        \end{tabular}
        }
    \end{minipage}%
    \quad
    \begin{minipage}[t]{0.47\textwidth}
        \centering
        \small
        \caption{Ablation on different re-indexing strategies on offline benchmark VideoMME. The gray row represents our default setting in all evaluations for offline benchmarks.}
        \label{tab:pos_ablation_offline}
        \resizebox{\linewidth}{!}{
        \begin{tabular}{lc|cccc}
        \toprule
         \multirow{2}{*}{\textbf{Model}} & \multirow{2}{*}{\textbf{Re-Indexing}} & \multicolumn{4}{c}{\textbf{VideoMME}} \\
        & & Short & Medium & Long & \textbf{Avg.}\\
        \midrule
        \llava & - & 69.89 & 55.11 & 48.00 & 57.67 \\
        \hspace{3pt}+ HERMES & lazy & 69.67 & 51.67 & 43.44 & 54.93 \\
        \rowcolor{gray!40} \hspace{3pt}+ HERMES & eager & \textbf{71.33} & 54.89 & \textbf{49.11} & \textbf{58.44} \\
        \bottomrule
        \end{tabular}
        }
    \end{minipage}
\end{table}
\paragraph{Position Re-Indexing Strategies}
For all streaming evaluations, we adopt the lazy position re-indexing strategy, while we use the eager re-indexing strategy for offline evaluations. Ablation studies in~\cref{tab:pos_ablation_streaming} and~\cref{tab:pos_ablation_offline} show the effectiveness of these strategies in their respective scenarios.

\paragraph{Guidance Prompts}
\begin{table*}[htbp]
    \centering
    \small
    \footnotesize
    \setlength{\tabcolsep}{3pt}
    \caption{Ablation on guidance prompts. The "generic prompt" refers to the guidance prompt utilized in the paper.}
    \label{tab:ablation_guidance_prompt}
    \vspace{2mm}
    \begin{tabular}{lcccccccc}
        \toprule
        \multirow{2}{*}{\textbf{Guidance Prompt}} & \textbf{StreamingBench} & \multicolumn{3}{c}{\textbf{OVO-Bench}} & \multicolumn{4}{c}{\textbf{VideoMME}} \\
        \cmidrule(lr){2-2} \cmidrule(lr){3-5} \cmidrule(lr){6-9}
        & Real-Time & Real-Time & Backward & \textbf{Avg.} & Short & Medium & Long & \textbf{Avg.} \\
        \midrule
        \multicolumn{9}{c}{\textbf{HERMES based on LLaVA-OV-7B}} \\
        \midrule
        generic prompt & 72.63 & 65.07 & 48.80 & 56.94 & \textbf{71.33} & \textbf{54.89} & \textbf{49.11} & \textbf{58.44} \\
        "What happens in the video?" & \textbf{72.75} & \textbf{65.49} & 48.60 & 57.05 & 71.11 & 54.11 & 47.67 & 57.63 \\
        "Describe the video in detail." & 72.71 & 65.39 & \textbf{49.48} & \textbf{57.44} & 70.33 & 53.44 & 47.78 & 57.19 \\
        "Summarize the content of the video." & 72.55 & 65.45 & 49.19 & 57.32 & \textbf{71.33} & 53.00 & 48.22 & 57.52 \\
        \midrule
        \multicolumn{9}{c}{\textbf{HERMES based on Qwen2.5-VL-7B}} \\
        \midrule
        generic prompt & 78.72 & 68.42 & 48.10 & 58.26 & \textbf{70.44} & \textbf{61.11} & \textbf{54.44} & \textbf{62.00} \\
        "What happens in the video?" & 78.84 & \textbf{69.40} & \textbf{49.36} & \textbf{59.38} & 70.11 & 59.33 & 53.22 & 60.89 \\
        "Describe the video in detail." & 78.92 & 68.90 & 49.13 & 59.02 & 70.33 & 59.78 & 53.78 & 61.30 \\
        "Summarize the content of the video." & \textbf{79.00} & 68.95 & 49.14 & 59.05 & 70.33 & 59.67 & \textbf{54.44} & 61.48 \\
        \bottomrule
    \end{tabular}
\end{table*}
To verify that the effectiveness of the token eviction strategy does not depend on a specific prompt design, we conduct an ablation study using three alternative guidance prompts. The results in~\cref{tab:ablation_guidance_prompt} show consistent performance across prompt variations, indicating that the method is largely insensitive to the exact wording or design of the guidance prompt.

\paragraph{Summary Tokens in Deep Layers}
In~\cref{sec: smoothing}, we aggregate the evicted tokens in each deep layer into one summary token at each compression step. The results in~\cref{tab:summary_token_ablation} indicate that these summary tokens effectively preserve long-term memory, leading to improved performance on VideoMME.
\section{Related Work}
\label{sec:related}

\paragraph{Streaming Video Understanding}
Existing MLLMs~\cite{gemini25, li2024llavaonevisioneasyvisualtask,bai2025qwen25vltechnicalreport, bai2025qwen3vltechnicalreport} are primarily designed for pre-defined offline videos and struggle with continuous streaming videos. While some prior works have adapted existing offline MLLMs to online settings~\cite{ timechatonline,zeng2025streamforestefficientonlinevideo,xu2025streamingvlmrealtimeunderstandinginfinite}, they rely on costly model-specific training. Training-free streaming methods, such as ReKV~\cite{di2025streamingvideoquestionansweringincontext} and LiveVLM~\cite{ning2025livevlmefficientonlinevideo}, prefill offload KV cache to external devices. At user query time, they retrieve the full KV cache and reconstruct it on the GPU, incurring high latency and overall memory usage. In contrast, StreamMem~\cite{yang2025streammemqueryagnostickvcache} heuristically reuses KV cache, but lacks fine-grained KV cache management and interpretability. Unlike prior training-free methods, \hermes is grounded in a systematic attention analysis with improved interpretability and reliability.

\paragraph{KV Cache Compression for Video Input}
Numerous KV cache compression techniques have been proposed for offline video understanding~\cite{ yang2024visionziplongerbetternecessary, wang2024dynamicvlmsimpledynamicvisual,wang2025videotreeadaptivetreebasedvideo, tao2025dycokedynamiccompressiontokens}, but most of these methods are poorly suited for streaming scenarios due to the unpredictable future frames and user queries~\cite{chen2025streamingtomstreamingtokencompression}.
Existing online KV cache compression paradigms~\cite{di2025streamingvideoquestionansweringincontext, ning2025livevlmefficientonlinevideo, yang2025streammemqueryagnostickvcache, chen2025streamingtomstreamingtokencompression} largely overlook the inherently hierarchical storage structure of the KV cache. \hermes addresses this gap by introducing a hierarchical KV cache management strategy, which enables fine-grained memory utilization and low-latency responses.

\section{Conclusion}
\label{sec:conclusion}

This paper proposes \hermes, a training-free framework for efficient streaming video understanding. Guided by mechanistic attention analysis, we conceptualizes KV cache as a hierarchical video memory system across multiple granularities. By introducing a cross-layer memory smoothing and position re-indexing, \hermes further enhances the understanding performance for long streaming input. Extensive experiments demonstrate that \hermes delivers accurate performance under continuously growing video streams, while consistently maintaining extremely low response latency and compact GPU memory usage, making it well suited for real-world streaming deployment.
\section*{Limitations}
While our evaluations have spanned a diverse range of MLLMs, due to computation resource constraints, we are unable to implement experiments on the 72B variant (e.g., Qwen2.5-VL-72B). Additionally, we do not investigate the integration of our method with other orthogonal training-free techniques, which may further enhance both understanding performance and efficiency of MLLMs in streaming video scenarios. We plan to conduct more extensive validation involving larger-scale MLLMs as computational overhead permits.
\section*{Acknowledgment}
This work was supported by the National Natural Science Foundation of China (No. U24B20181 and 62521004).
This research/project is supported by the National Research Foundation, Singapore under its National Large Language Models Funding Initiative (AISG Award No: AISG-NMLP-2024-002). Any opinions, findings and conclusions or recommendations expressed in this material are those of the author(s) and do not reflect the views of National Research Foundation, Singapore.

\clearpage
\bibliographystyle{plainnat}
\bibliography{main,datasets}

\clearpage
\beginappendix

\startcontents[app]
\begingroup
  \renewcommand{\contentsname}{Appendix Contents}
  \section*{\contentsname}
  \printcontents[app]{}{1}{}
\endgroup
\newpage

\crefalias{section}{appendix}
\crefalias{subsection}{appendix}





\section{More Attention Visualization}
\begin{figure*}[htbp]
\centering

\begin{subfigure}{0.95\textwidth}
    \centering
    \includegraphics[width=\textwidth]{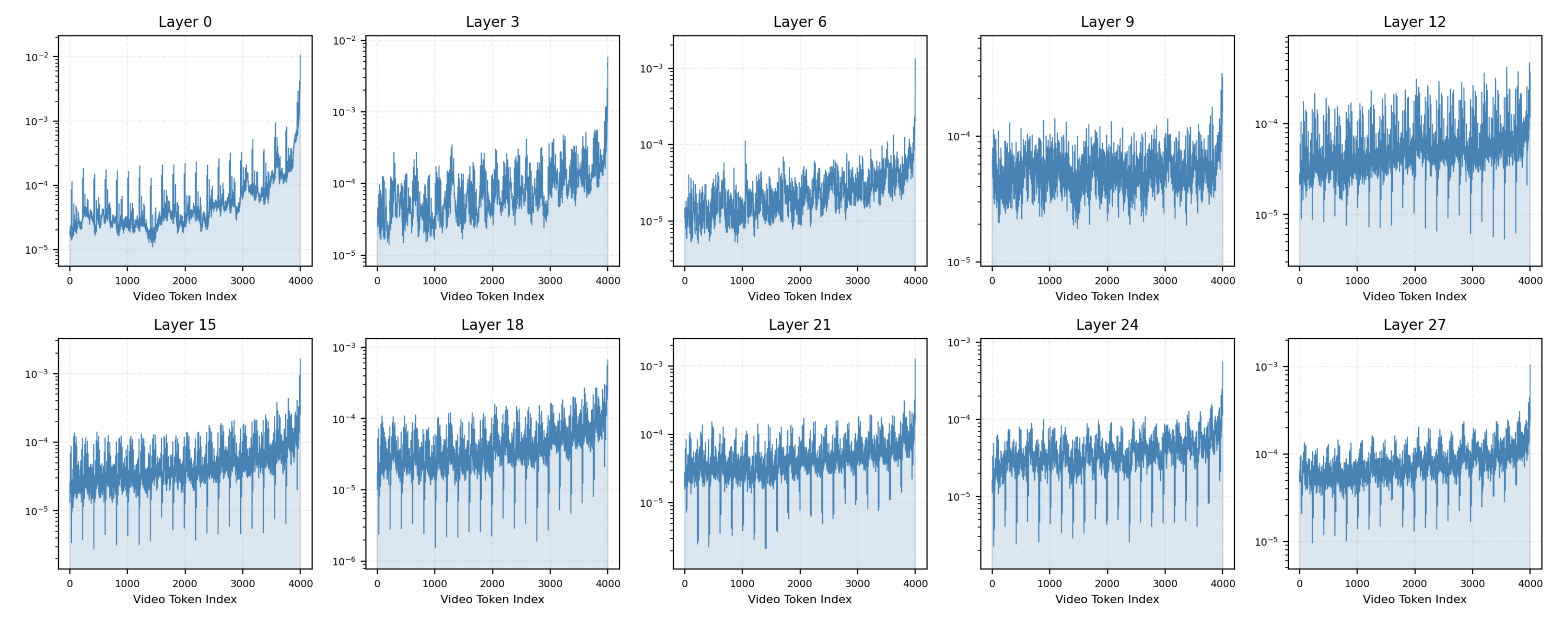}
    \caption{Sliding window of 4,000 video tokens}
\end{subfigure}

\vspace{0.8em}

\begin{subfigure}{0.95\textwidth}
    \centering
    \includegraphics[width=\textwidth]{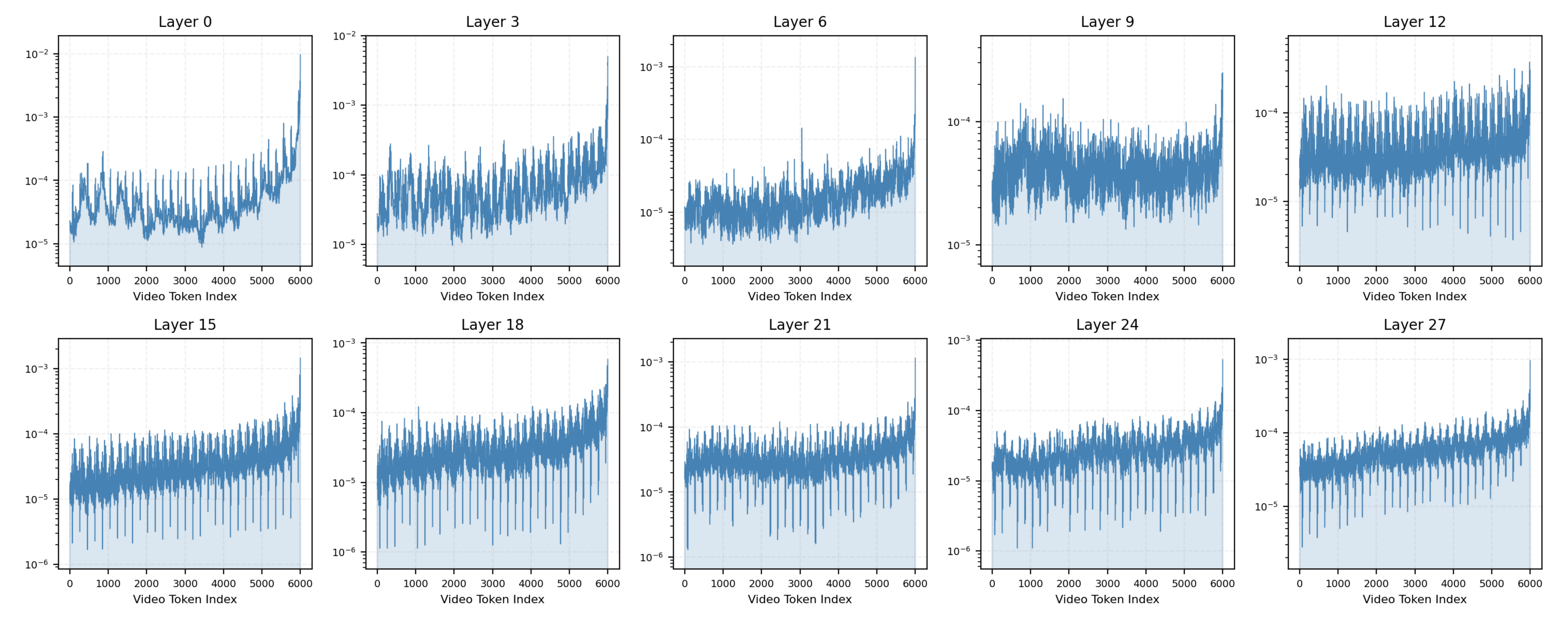}
    \caption{Sliding window of 6,000 video tokens}
\end{subfigure}

\vspace{0.8em}

\begin{subfigure}{0.95\textwidth}
    \centering
    \includegraphics[width=\textwidth]{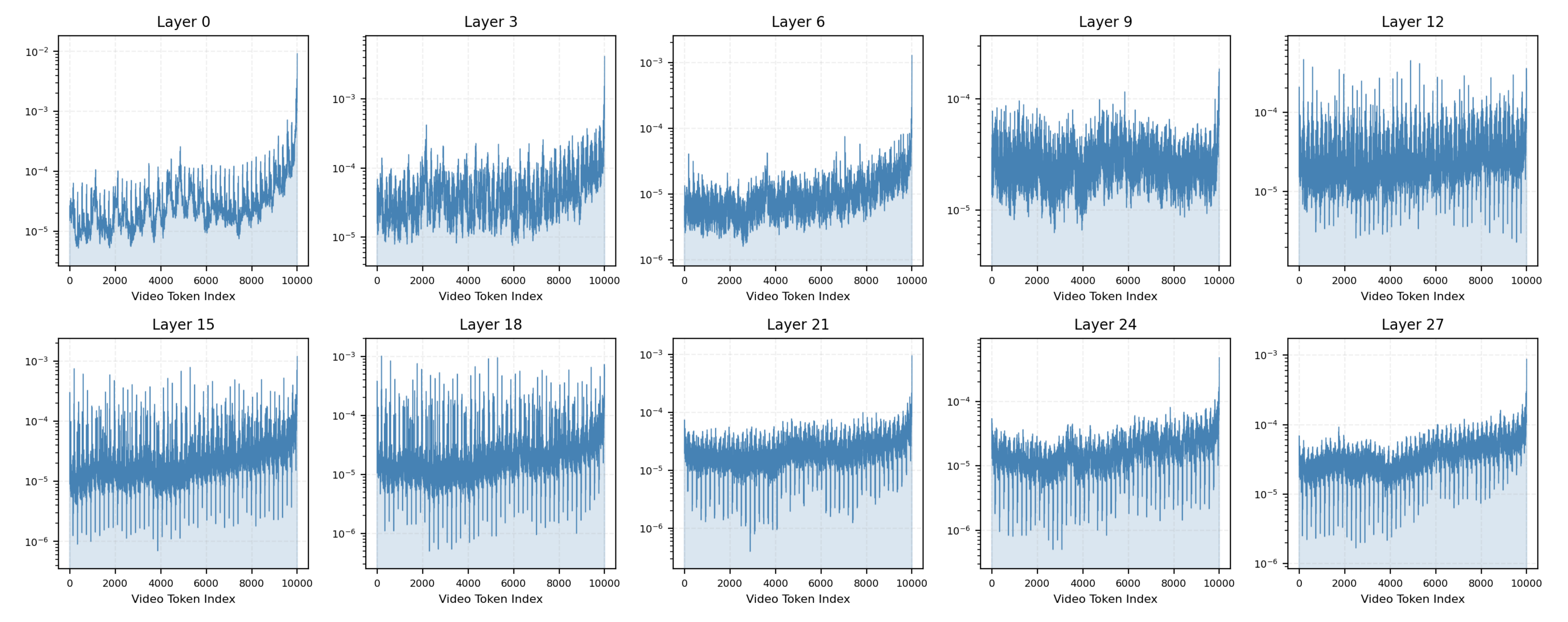}
    \caption{Sliding window of 10,000 video tokens}
\end{subfigure}

\caption{
Visualization of the average attention weights of video tokens in \llava under different sliding window sizes.
}
\label{fig:more_vis}
\end{figure*}

We provide more detailed attention visualization in~\cref{fig:more_vis} under different sliding window sizes, showing that the observed attention patterns consistently hold across varying window lengths, thus confirming the generality of the findings in~\cref{sec: investigation}.

\label{app:attn_vis}

\section{Guidance Prompt}
\label{app:prompt}
The following two figures show the local and global guidance prompt with and without conversation history to guide the token compression, respectively. For the deep layers, since they primarily focus on frame-level global semantic information, we employ a global guidance prompt as a pseudo-query to extract attention weights of video tokens. In contrast, the middle layers lie in a transition between recency-biased attention and global semantic focus. Therefore, we adopt a hybrid guidance strategy, in which the local guidance prompt and the global guidance prompt are concatenated into a single prompt string to jointly guide the token compression.

\begin{figure}[ht]
    \centering
    \begin{minipage}[t]{0.47\textwidth}
    \begin{tcolorbox}[colback=black!7.5!white, colframe=black!30!white, fontupper=\footnotesize, fonttitle=\footnotesize]
    Find recent details related to: \{last\_conv\}. Describe the current scene in detail, focusing on specific objects, fine-grained actions, and spatial relationships.
    \end{tcolorbox}
    \caption{Local guidance prompt to guide the token compression if conversation history exists. "last\_conv" refers to the last user query and the corresponding model answer from the conversation history.}
    \end{minipage}
    \quad
    \begin{minipage}[t]{0.47\textwidth}
    \begin{tcolorbox}[colback=black!7.5!white, colframe=black!30!white, fontupper=\footnotesize, fonttitle=\footnotesize]
    Describe the current scene in detail, focusing on specific objects, fine-grained actions, and spatial relationships.
    \end{tcolorbox}
    \caption{Local guidance prompt to guide the token compression if there is no conversation history.}
    \end{minipage}
\end{figure}

\begin{figure}[ht]
    \centering
    \begin{minipage}[t]{0.47\textwidth}
    \begin{tcolorbox}[colback=black!7.5!white, colframe=black!30!white, fontupper=\footnotesize, fonttitle=\footnotesize]
    Context summary: \{last\_conv\}. Summarize the video narrative, identifying main characters, key events, timeline changes, and the overall theme.
    \end{tcolorbox}
    \caption{Global guidance prompt to guide the token compression if conversation history exists. "last\_conv" refers to the last user query and the corresponding model answer from the conversation history.}
    \end{minipage}
    \quad
    \begin{minipage}[t]{0.47\textwidth}
    \begin{tcolorbox}[colback=black!7.5!white, colframe=black!30!white, fontupper=\footnotesize, fonttitle=\footnotesize]
    Summarize the video narrative, identifying main characters, key events, timeline changes, and the overall theme.
    \end{tcolorbox}
    \caption{Global guidance prompt to guide the token compression if there is no conversation history.}
    \end{minipage}
\end{figure}
\section{Configuration of Cross-Layer Memory Smoothing}
\label{app:smooth_config}
Given that long-term memory tends to remain relatively stable, while short-term memory focuses on diverse perception, we set different $\lambda$ for different layer stages:
\begin{equation}
\lambda_l =
\begin{cases}
0.1,   & \text{if } l \in \mathcal{L}_{shallow} \\
0.3,   & \text{if } l \in \mathcal{L}_{middle} \\
0.4, & \text{if } l \in \mathcal{L}_{deep}
\end{cases}
\end{equation}
The ablation study~\cref{tab:lambda_ablation} shows the effectiveness of this hyperparameter choice.
\section{Details of evaluated benchmarks}
\label{app:details_of_benchmarks}

\begin{table}[h]
\centering
    \begin{minipage}[h]{0.49\textwidth}
        \centering
        \caption{\textbf{Key statistics of the streaming benchmarks.} In the “Type” column, "MC" denotes multiple-choice questions, while "OE" denotes open-ended questions. In the "Benchmark" column, "rt" denotes real-time understanding subset, while "bw" denotes backward tracing subset.}
        \label{tab:streaming_bmk}
        \small
        \resizebox{\linewidth}{!}{
        \begin{tabular}{lcrrc}
            \toprule
            \textbf{Benchmark} & \textbf{Duration} & \textbf{\#Videos} & \textbf{\#QA} & \textbf{Type} \\
            \midrule
            StreamingBench$_{\text{rt}}$ & 10.1min & 500 & 2,500 & MC \\
            OVO-Bench$_{\text{bw}}$ & 5.9 min & 275 & 631 & MC \\
            OVO-Bench$_{\text{rt}}$ & 8.8 min & 237 & 837 & MC \\
            RVS-Ego & 60 min & 10 & 1,465 & OE \\
            RVS-Movie & 30 min & 22 & 1,905 & OE \\
            \bottomrule
        \end{tabular}
        }
    \end{minipage}
    \hfill%
    \begin{minipage}[h]{0.49\textwidth}
        \centering
        \small
        \caption{\textbf{Key statistics of the offline benchmarks.} In the "Type" column, "MC" denotes multiple-choice questions.}
        \label{tab:offline_bmk}
        \resizebox{\linewidth}{!}{
        \begin{tabular}{lcrrc}
            \toprule
            \textbf{Benchmark} & \textbf{Duration} & \textbf{\#Videos} & \textbf{\#QA} & \textbf{Type} \\
            \midrule
            MVBench & 16 s & 3,641 & 4,000 & MC \\
            Egoschema & 3 min & 5,063 & 5,063 & MC \\
            VideoMME & 17 min & 900 & 2,700 & MC \\
            \bottomrule
        \end{tabular}
        }
    \end{minipage}
\end{table}
\subsection{Streaming Benchmarks}

\begin{itemize} [leftmargin=*]
\itemsep0em 
\item \textbf{StreamingBench}~\cite{lin2024streamingbenchassessinggapmllms} assesses the streaming video understanding capabilities of MLLMs. It evaluates three core aspects: real-time visual understanding, omni-source understanding, and contextual understanding. The Real-Time Visual Understanding subset is the most extensive component, featuring 2,500 questions across 500 videos. It covers 10 tasks, such as object perception and causal reasoning. In this paper, we focus on the Real-Time Visual Understanding subset for evaluation.

\item
\textbf{OVO-Bench}~\cite{li2025ovobenchfarvideollmsrealworld} evaluates the online reasoning and temporal awareness of MLLMs, featuring 644 videos with approximately 2,800 fine-grained multiple-choice QA pairs. It organizes 12 tasks into three distinct categories, which are real-time visual perception, backward tracing, and forward active responding. Given that we do not focus on the proactive responding ability of MLLMs in this paper, we exclusively utilize the real-time perception and the backward tracing subsets.



\item
\textbf{RVS-Ego} and \textbf{RVS-Movie}~\cite{zhang2024flashvstreammemorybasedrealtimeunderstanding} are designed to evaluate the real-time understanding capabilities of models in online streaming scenarios. The datasets consist of 10 long ego-centric videos from the Ego4D dataset~\cite{grauman2022ego4dworld3000hours} and 22 long movie clips from the MovieNet dataset~\cite{huang2020movienetholisticdatasetmovie} dataset, totaling over 21 hours of video content.
\end{itemize}

\subsection{Offline Benchmarks}
\begin{itemize}[leftmargin=*]
\itemsep0em
\item
\textbf{MVBench}~\cite{li2024mvbenchcomprehensivemultimodalvideo} systematically evaluates the temporal understanding capabilities of MLLMs. It utilizes a novel static-to-dynamic method to define 20 distinct temporal tasks, such as action sequence and moving direction, which cannot be effectively solved with a single frame. The videos are collected from a wide range of datasets, including NTU RGB+D~\cite{shahroudy2016nturgbdlargescale}, Perception~\cite{pătrăucean2023perceptiontestdiagnosticbenchmark}, etc.

\item \textbf{Egoschema}~\cite{mangalam2023egoschemadiagnosticbenchmarklongform} is a diagnostic benchmark designed to assess long-form video understanding abilities. Derived from Ego4D~\cite{grauman2022ego4dworld3000hours}, it consists of over 5,000 human-curated multiple-choice QA pairs associated with egocentric video clips.

\item \textbf{VideoMME}~\cite{fu2025videommefirstevercomprehensiveevaluation} is a full-spectrum, multimodal benchmark designed for the comprehensive evaluation of MLLMs in video analysis. It comprises 900 manually curated videos spanning six primary domains and diverse durations to assess temporal adaptability. The dataset features 2,700 high-quality QA pairs that necessitate processing multimodal inputs, including video frames, subtitles, and audio.
\end{itemize}
\section{Details of Position Re-Indexing}

Inspired by StreamingVLM's strategy of managing positional stability in streaming scenarios~\cite{xu2025streamingvlmrealtimeunderstandinginfinite},  we adopt a unified left-compaction re-indexing scheme to eliminate positional gaps introduced by KV-cache pruning while preserving the semantic anchoring of the system prompt. Concretely, system text tokens are kept fixed to provide a stable textual anchor, whereas retained video tokens are re-indexed in a left-compact manner and placed contiguously after the static prefix. To reuse cached key states without re-computation, we further apply a delta-based rotary correction that compensates for the positional displacement.

\subsection{Re-indexing for LLaVA-OV (1D RoPE)}
\label{app:1d_pos}

LLaVA-OV employs standard 1D RoPE, where each token is associated with a scalar positional index $p$. Therefore, we perform left-compaction of the 1D indices: the system prefix positions remain unchanged, while the retained positions of video tokens are reassigned to form a dense contiguous segment immediately following the fixed prefix.

Let \texttt{offset} denote the length of the system prompt prefix tokens, and let
\[
\mathcal{P} = \{ p_0 < p_1 < \cdots < p_{N-1} \}
\]
be the sorted set of retained video token positions (excluding the fixed prefix). For a retained video token originally at position $p_{\mathrm{old}} \in \mathcal{P}$, its compacted 1D position is defined as
\begin{equation}
p_{\mathrm{new}}
=
\texttt{offset}
+
\operatorname{rank}_{\mathcal{P}}\!\left(p_{\mathrm{old}}\right).
\end{equation}
This mapping removes gaps while preserving the original temporal ordering along the stream, and ensures that the video region occupies a dense range directly after the static text region.

To align cached key states with the updated positions, we avoid re-generating keys and instead apply a rotary delta correction induced by the positional shift. For a cached key vector $\mathbf{k}_{\mathrm{old}}$ associated with position $p_{\mathrm{old}}$ and remapped to $p_{\mathrm{new}}$, we compute
\begin{equation}
\mathbf{k}_{\mathrm{new}}
=
\mathbf{k}_{\mathrm{old}}
\odot
\mathrm{RotaryDelta}\!\left(p_{\mathrm{old}}, p_{\mathrm{new}}\right),
\end{equation}
where the relative phase shift is
\begin{equation}
\mathrm{RotaryDelta}\!\left(p_{\mathrm{old}}, p_{\mathrm{new}}\right)
=
e^{i(p_{\mathrm{new}} - p_{\mathrm{old}})\boldsymbol{\theta}},
\end{equation}
and $\boldsymbol{\theta}$ denotes the RoPE frequency vector. This update preserves the correctness of attention under the new indexing while enabling direct reuse of the cached KV states.

\subsection{Re-indexing for Qwen2.5-VL (3D M-RoPE)}
\label{app:3d_pos}

For Qwen2.5-VL, video tokens are indexed by a 3D M-RoPE coordinate $\mathbf{p} = (p^{(t)}, p^{(h)}, p^{(w)})$, covering temporal and spatial dimensions. After pruning, the retained video tokens typically occupy sparse coordinates along each dimension $d \in \{t, h, w\}$. To eliminate the gaps without disturbing the monotonic ordering, we apply dimension-wise left-compaction independently along each axis, while keeping the system token prefix fixed.

Let
\[
\mathcal{P}^{(d)} = \{ p^{(d)}_0 < p^{(d)}_1 < \cdots < p^{(d)}_{N_{d}-1} \}
\]
denote the sorted set of retained coordinates along dimension $d$. For a token originally located at $p^{(d)}_{\mathrm{old}} \in \mathcal{P}^{(d)}$, its compacted coordinate is defined by its rank within $\mathcal{P}^{(d)}$, shifted by the fixed prefix \texttt{offset}:
\begin{equation}
p^{(d)}_{\mathrm{new}}
=
\texttt{offset}
+
\operatorname{rank}_{\mathcal{P}^{(d)}}\!\left(p^{(d)}_{\mathrm{old}}\right),
\qquad
d \in \{t, h, w\}.
\end{equation}
This procedure yields a dense and contiguous $(t,h,w)$ grid for the video tokens placed immediately after the static text region, thereby ensuring positional continuity while preserving the distinct semantic roles of temporal and spatial indices.

As in the 1D case, we reuse cached keys by applying a M-RoPE correction. Given a key $\mathbf{k}_{\mathrm{old}}$ associated with
\[
\mathbf{p}_{\mathrm{old}} = (p^{(t)}_{\mathrm{old}}, p^{(h)}_{\mathrm{old}}, p^{(w)}_{\mathrm{old}})
\]
and remapped to
\[
\mathbf{p}_{\mathrm{new}} = (p^{(t)}_{\mathrm{new}}, p^{(h)}_{\mathrm{new}}, p^{(w)}_{\mathrm{new}}),
\]
the corrected key is obtained as
\begin{equation}
\mathbf{k}_{\mathrm{new}}
=
\mathbf{k}_{\mathrm{old}}
\odot
\mathrm{RotaryDelta}\!\left(\mathbf{p}_{\mathrm{old}}, \mathbf{p}_{\mathrm{new}}\right),
\end{equation}
with the relative phase shift:
\begin{equation}
\mathrm{RotaryDelta}\!\left(\mathbf{p}_{\mathrm{old}}, \mathbf{p}_{\mathrm{new}}\right)
=
\operatorname*{Concat}_{d \in \{t, h, w\}}
\left(
e^{i
(
p^{(d)}_{\mathrm{new}} - p^{(d)}_{\mathrm{old}}
)
\boldsymbol{\theta}^{(d)}
}\right),
\end{equation}
where $\operatorname*{Concat}$ denotes the concatenation operation along the channel dimension, and $\boldsymbol{\theta}^{(d)}$ represents the rotary frequency vector corresponding to the channel section allocated for dimension $d$.

\section{Algorithm of Summary Tokens}




\begin{algorithm}
\label{alg:summary}
\caption{Summary Token Aggregation}
\small
\begin{algorithmic}
\REQUIRE 
    $K_{p}, V_{p}$: Pruned KV tensors from visual tokens;
    $P_{p}$: Original position indices of pruned tokens;
    $t$: Target position index for the summary token. \\
\ENSURE
    $k_{sum}, v_{sum}$: Single aggregated summary token cache.\\

\item[]

\STATE \textbf{Step 1: Aggregate Value} \\
\# Simple spatial mean \\
\STATE $v_{sum} \leftarrow \text{Mean}(V_{p})$ 

\item[]

\STATE \textbf{Step 2: Aggregate Key} \\
\# Phase alignment before pooling \\
\STATE $\Delta\theta \leftarrow \text{RotaryDelta}(P_{p} \to t)$ \\
\# Calculate rotation shift from $P_p$ to $t$ \\
\STATE $K_{aligned} \leftarrow \text{ApplyDelta}(K_{p}, \Delta\theta)$ \\
\# Align all keys to the same phase \\
\STATE $k_{sum} \leftarrow \text{Mean}(K_{aligned})$

\item[]

\STATE \textbf{Step 3: Update KV Cache} \\
\STATE $K_{new} \leftarrow \text{Concat}([K_{kept}, k_{sum}])$
\STATE $V_{new} \leftarrow \text{Concat}([V_{kept}, v_{sum}])$

\item[]

\RETURN $K_{new}, V_{new}$
\end{algorithmic}
\end{algorithm}
\section{Full Performances}
\label{sec:full_performance}

\subsection{StreamingBench}
\begin{table*}[t]
    \centering
    \caption{\textbf{Accuracy comparison (\%) on StreamingBench focusing on \textit{Real-Time Visual Understanding} tasks}. Real-Time Visual Understanding tasks consists of Object Perception (OP), Causal Reasoning (CR), Clips Summarization (CS), Attribute Perception (ATP), Event Understanding (EU), Text-Rich Understanding (TR), Prospective Reasoning (PR), Spatial Understanding (SU), Action Perception (ACP), and Counting (CT).}
    \label{tab:streamingbench_full}
    \small
    \begin{adjustbox}{max width=\linewidth}
    \begin{tabular}{l c | c c c c c c c c c c | c}
    \toprule
    \multirow{1}{*}{\textbf{Model}} & \multirow{1}{*}{\textbf{\#Frames}} & \textbf{OP} & \textbf{CR} & \textbf{CS} & \textbf{ATP} & \textbf{EU} & \textbf{TR} & \textbf{PR} & \textbf{SU} & \textbf{ACP} & \textbf{CT} & \textbf{Avg.} \\
    \midrule
    \midrule
    Human & - & 89.47 & 92.00 & 93.60 & 91.47 & 95.65 & 92.52 & 88.00 & 88.75 & 89.74 & 91.30 & 91.46 \\
    \midrule
    \multicolumn{13}{c}{\textbf{Proprietary MLLMs}} \\
    \midrule
    Gemini 1.5 pro~\cite{gemini25} & 1 fps & 79.02 & 80.47 & 83.54 & 79.67 & 80.00 & 84.74 & 77.78 & 64.23 & 71.95 & 48.70 & 75.69 \\
    GPT-4o~\cite{openai2024gpt4ocard} & 64 
        & 77.11 & 80.47 & 83.91 & 76.47 & 70.19 & 83.80 & 66.67 & 62.19 & 69.12 & 49.22 & 73.28 \\
    Claude 3.5 Sonnet~\cite{claude3_5} & 20 
        & 73.33 & 80.47 & 84.09 & 82.02 & 75.39 & 79.53 & 61.11 & 61.79 & 69.32 & 43.09 & 72.44 \\
    \midrule
    \multicolumn{13}{c}{\textbf{Open-source Offline MLLMs}} \\
    \midrule
    Video-LLaMA2-7B~\cite{videollama2} & 32 & 
            55.86 & 55.47 & 57.41 & 58.17 & 52.80 & 43.61 & 39.81 & 42.68 & 45.61 & 35.23 & 49.52 \\
    VILA-1.5-8B~\cite{vila} & 14 &  
        53.68 & 49.22 & 70.98 & 56.86 & 53.42 & 53.89 & 54.63 & 48.78 & 50.14 & 17.62 & 52.32 \\
    Video-CCAM-14B~\cite{videoccam} & 96 & 
        56.40 & 57.81 & 65.30 & 62.75 & 64.60 & 51.40 & 42.59 & 47.97 & 49.58 & 31.61 & 53.96 \\
    LongVA-7B~\cite{longva} & 128 & 
        70.03 & 63.28 & 61.20 & 70.92 & 62.73 & 59.50 & 61.11 & 53.66 & 54.67 & 34.72 & 59.96 \\
    InternVL-V2-8B~\cite{internvl2} & 16 & 
        68.12 & 60.94 & 69.40 & 77.12 & 67.70 & 62.93 & 59.26 & 53.25&  54.96 & 56.48 & 63.72 \\
    Kangaroo-7B~\cite{kangaroo} & 64 & 
        71.12 & 84.38 & 70.66 & 73.20 & 67.08 & 61.68 & 56.48 & 55.69 & 62.04 & 38.86 & 64.60 \\
    LLaVA-NeXT-Video-32B~\cite{llava-next} & 64 & 
        78.20 & 70.31 & 73.82 & 76.80 & 63.35 & 69.78 & 57.41 & 56.10&  64.31 & 38.86 & 66.96 \\
    MiniCPM-V-2.6-8B~\cite{minicpm} & 32 &  
        71.93 & 71.09 & 77.92 & 75.82 & 64.60 & 65.73 & 70.37 & 56.10 & 62.32 & 53.37 & 67.44 \\
    \addlinespace[1pt]
    
    \midrule
    \multicolumn{13}{c}{\textbf{Open-source  Online MLLMs}} \\
    \midrule
    Flash-VStream-7B~\cite{flashvstream} & - 
        & 25.89 & 43.57 & 24.91 & 23.87 & 27.33 & 13.08 & 18.52 & 25.20 & 23.87 & 48.70 & 23.23 \\
    VideoLLM-online-8B~\cite{videollmonline} & 2 fps 
        & 39.07 & 40.06 & 34.49 & 31.05 & 45.96 & 32.40 & 31.48 & 34.16 & 42.49 & 27.89 & 35.99 \\
    Dispider-7B~\cite{dispider} & 1 fps 
        & 74.92 & 75.53 & 74.10 & 73.08 & 74.44 & 59.92 & 76.14 & 62.91 & 62.16 & 45.80 & 67.63 \\
    TimeChat-Online-7B~\cite{timechatonline} &1 fps
        & 80.22 & 82.03 & 79.50 & 83.33 & 76.10 & 78.50 & 78.70 & 64.63 & 69.60 & 57.98 & 75.36 \\
    StreamForest-7B~\cite{zeng2025streamforestefficientonlinevideo} &1 fps
        & 83.11 & 82.81 & 82.65 & 84.26 & 77.50 & 78.19 & 76.85 & 69.11 & 75.64 & 54.40 & 77.26 \\

    \midrule
    \multicolumn{13}{c}{\textbf{Training-free Offline-to-Online Methods}} \\
    \midrule
    LLaVA-OV-7B~\cite{li2024llavaonevisioneasyvisualtask} & 64  &  78.75 & 78.12 & 80.76 & \textbf{81.19} & 71.70 & 72.59 & 72.22 & 63.82&  66.01 & 38.34 & 71.34 \\

    \hspace{3pt} + ReKV~\cite{di2025streamingvideoquestionansweringincontext} & 0.5 fps &  76.02 & 81.25 & 77.92 & 76.90 & 66.04 & 66.04 & 69.44 & 60.98& 64.31 & \textbf{49.22} & 69.22 \\

    \hspace{3pt} + LiveVLM~\cite{ning2025livevlmefficientonlinevideo} & 0.5 fps & \textbf{81.47} & 78.13 & 83.28 & 79.08 & 69.57 & \textbf{74.14} & \textbf{75.00} & \textbf{69.11} &  67.71 & 40.41 & 72.92 \\

    \hspace{3pt} + StreamKV~\cite{chen2025streamkvstreamingvideoquestionanswering} & 0.5 fps & 73.80 & 77.30 & 85.90 & 77.50 & \textbf{73.30} & 63.90 & 69.40 & 61.40 &  63.20 & 35.80 & 68.80 \\

    \rowcolor{gray!20} \hspace{3pt} + HERMES (6K tokens) & 0.5 fps & 77.93 & \textbf{82.03} & 86.12 & \textbf{81.19} & 66.04 & 73.52 & 74.07 & 63.01 &  67.71 & 45.08 & 72.63 \\

     \rowcolor{gray!40} \hspace{3pt} + HERMES (4K tokens) & 0.5 fps & 79.02 & 81.25 & \textbf{87.70} & 80.20 & 69.18 & 71.96 & 73.15 & 66.26 &  \textbf{69.41} & 43.52 & \textbf{73.23} \\

    \midrule

    LLaVA-OV-0.5B~\cite{li2024llavaonevisioneasyvisualtask} & 64  &  71.39 & 57.81 & 65.93 & 69.64 & 69.18 & 55.76 & 57.41 & \textbf{52.85} & 62.04 & 16.58 & 59.64 \\

    \hspace{3pt} + ReKV~\cite{di2025streamingvideoquestionansweringincontext} & 0.5 fps &  65.12 & 60.16 & 66.56 & 66.01 & 66.67 & 52.96 & 57.41 & 48.37&  60.34 & 18.13 & 57.39 \\

    \rowcolor{gray!20} \hspace{3pt} + HERMES (6K tokens) & 0.5 fps & 71.93 & 60.16 & 69.09 & 71.29 & 68.55 & 57.32 & \textbf{60.19} & 51.22 &  \textbf{63.74} & \textbf{19.69} & 61.04 \\

     \rowcolor{gray!40} \hspace{3pt} + HERMES (4K tokens) & 0.5 fps & \textbf{72.21} & \textbf{61.72} & \textbf{70.98} & \textbf{72.94} & \textbf{72.33} & \textbf{57.94} & \textbf{60.19} & \textbf{52.85} &  \textbf{63.74}& 19.17 & \textbf{62.04} \\

    \midrule    
    Qwen2.5-VL-7B~\cite{bai2025qwen25vltechnicalreport} & 1 fps & 
        77.93 & 76.56 & 78.55 & 80.86 & 76.73 & 76.95 & 80.56 & 65.45 & 65.72 & \textbf{52.85} & 73.31 \\
    \rowcolor{gray!20} \hspace{3pt} + HERMES (6K tokens) & 1 fps & 83.38 & 78.91 & 86.12 & 87.13 & \textbf{78.62} & \textbf{86.60} & \textbf{84.26} & 74.80 &  71.39 & 46.63 & 78.72 \\
        
    \rowcolor{gray!40} \hspace{3pt} + HERMES (4K tokens) & 1 fps & \textbf{83.65} & \textbf{81.25} & \textbf{88.01} & \textbf{87.46} & 76.73 & \textbf{86.60} & 82.41 & \textbf{76.02} &  \textbf{73.94} & 46.63 & \textbf{79.44} \\

    \midrule
    
    Qwen2.5-VL-32B~\cite{bai2025qwen25vltechnicalreport} & 1 fps & 
        76.29 & 79.69 & 78.55 & \textbf{83.50} & 76.10 & 79.44 & 80.56 & 61.38 & 68.27 & \textbf{59.07} & 74.27 \\
    \rowcolor{gray!20} \hspace{3pt} + HERMES (6K tokens) & 1 fps & \textbf{84.47} & 79.69 & \textbf{87.70} & 83.17 & \textbf{81.76} & \textbf{88.16} & 86.11 & 74.80 &  \textbf{77.62} & 49.22 & \textbf{80.20} \\
        
    \rowcolor{gray!40} \hspace{3pt} + HERMES (4K tokens) & 1 fps & 83.92 & \textbf{80.47} & \textbf{87.70} & \textbf{83.50} & 80.50 & \textbf{88.16} & \textbf{87.04} & \textbf{75.20} &  77.34 & 48.19 & 80.08 \\

    \midrule
    
    Qwen3-VL-8B~\cite{bai2025qwen3vltechnicalreport} & 2 fps & 
        \textbf{85.83} & \textbf{79.69} & 85.49 & 86.80 & \textbf{79.87} & \textbf{89.10} & 82.41 & 66.26 & 68.84 & \textbf{56.99} & 78.92 \\
    \rowcolor{gray!20} \hspace{3pt} + HERMES (6K tokens) & 2 fps & 84.47 & 78.91 & \textbf{90.54} & \textbf{88.78} & 78.62 & 88.79 & \textbf{87.96} & 74.39 & \textbf{79.32} & 48.70 & \textbf{81.32} \\
        
    \rowcolor{gray!40} \hspace{3pt} + HERMES (4K tokens) & 2 fps & 84.47 & \textbf{79.69} & 89.91 & \textbf{88.78} & 76.73 & 88.16 & 87.04 & \textbf{76.83} & 79.04 & 49.22 & 81.28 \\

    \midrule
    
    Qwen3-VL-4B~\cite{bai2025qwen3vltechnicalreport} & 2 fps & 
        \textbf{84.20} & \textbf{81.25} & 84.23 & 85.15 & 77.36 & \textbf{87.54} & 79.63 & 68.29 & 70.25 & \textbf{56.99} & 78.32 \\
    \rowcolor{gray!20} \hspace{3pt} + HERMES (6K tokens) & 2 fps & 79.56 & 78.12 & \textbf{88.33} & 87.79 & 79.25 & 85.98 & \textbf{85.19} & 72.36 & \textbf{74.22} & 43.52 & \textbf{78.40} \\
        
    \rowcolor{gray!40} \hspace{3pt} + HERMES (4K tokens) & 2 fps & 79.84 & 77.34 & 88.01 & \textbf{88.12} & \textbf{79.87} & 85.67 & 84.26 & \textbf{73.58} & 73.65 & 41.45 & 78.24 \\

    \bottomrule
    \end{tabular}
\end{adjustbox}
\end{table*}

\subsection{OVO-Bench}
\begin{table*}[t]
    \small
    \centering
    \caption{\textbf{Accuracy comparison (\%) on OVO-Bench focusing on \textit{Real-Time Visual Perception} and \textit{Backward Tracing} tasks}. Real-Time Visual Perception tasks consist of Optical Character Recognition (OCR), Action Recognition (ACR), Attribute Recognition (ATR), Spatial Understanding (STU), Future Prediction (FPD), Object Recognition (OJR). Backward Tracing tasks consists of Episodic Memory (EPM), Action Sequence Identification (ASI), Hallucination Detection (HLD).}
    \label{tab:ovobench_full}

    \begin{adjustbox}{max width=\linewidth}
    \begin{tabular}{p{3.7cm}@{}c@{\hspace{3pt}}|cccccc|c|ccc|c|c}
    \toprule
    \multirow{2}{*}{\textbf{Model}} & \multirow{2}{*}{\textbf{\#Frames}} & 
    \multicolumn{7}{c|}{\textbf{Real-Time Visual Perception}} & 
    \multicolumn{4}{c|}{\textbf{Backward Tracing}} & 
    \multirow{2}{*}{\textbf{Overall}} \\
    \addlinespace[2pt]
    \cmidrule[0.5pt](lr){3-9} \cmidrule[0.5pt](lr){10-13} 
    \addlinespace[2pt]
    & & OCR & ACR & ATR & STU & FPD & OJR & Avg. & 
    EPM & ASI & HLD & Avg. & Avg. \\
    \midrule
    \midrule
    Human & - & 93.96 & 92.57 & 94.83 & 92.70 & 91.09 & 94.02 & 93.20 & 92.59 & 93.02 & 91.37 & 92.33 & 92.77 \\
    \midrule
    \multicolumn{14}{c}{\textbf{Proprietary MLLMs}} \\
    \midrule
    Gemini 1.5 Pro~\cite{gemini25} & 1 fps & 85.91 & 66.97 & 79.31 & 58.43 & 63.37 & 61.96 & 69.32 & 58.59 & 76.35 & 52.64 & 62.54 & 65.93 \\
    GPT-4o~\cite{openai2024gpt4ocard} & 64 & 69.80 & 64.22 & 71.55 & 51.12 & 70.30 & 59.78 & 64.46 & 57.91 & 75.68 & 48.66 & 60.75 & 62.61 \\
    \midrule
    \multicolumn{14}{c}{\textbf{Open-source Offline MLLMs}} \\
    \midrule
    LLaVA-Video-7B~\cite{zhang2025llavavideovideoinstructiontuning} & 64 & 69.80 & 59.63 & 66.38 & 50.56 & 72.28 & 61.41 & 63.34 & 51.18 & 64.19 & 9.68 & 41.68 & 52.51 \\
    Qwen2-VL-7B~\cite{wang2024qwen2vlenhancingvisionlanguagemodels} & 64 & 69.13 & 53.21 & 63.79 & 50.56 & 66.34 & 60.87 & 60.65 & 44.44 & 66.89 & 34.41 & 48.58 & 54.62 \\
    InternVL2-8B~\cite{internvl2} & 64 & 68.46 & 58.72 & 68.97 & 44.94 & 67.33 & 55.98 & 60.73 & 43.10 & 61.49 & 27.41 & 44.00 & 52.37 \\
    LongVU-7B~\cite{shen2024longvuspatiotemporaladaptivecompression} & 1 fps & 55.70 & 49.54 & 59.48 & 48.31 & 68.32 & 63.04 & 57.40 & 43.10 & 66.22 & 9.14 & 39.49 & 48.45 \\
    \midrule
    \multicolumn{14}{c}{\textbf{Open-source Online MLLMs}} \\
    \midrule
    VideoLLM-online-8B~\cite{videollmonline} & 2 fps & 8.05 & 23.85 & 12.07 & 14.04 & 45.54 & 21.20 & 20.79 & 22.22 & 18.80 & 12.18 & 17.73 & 19.26 \\
    Flash-VStream-7B~\cite{flashvstream} & 1 fps & 25.50 & 32.11 & 29.31 & 33.71 & 29.70 & 28.80 & 29.86 & 36.36 & 33.78 & 5.91 & 25.35 & 27.61 \\
    Dispider-7B~\cite{dispider} & 1 fps & 57.72 & 49.54 & 62.07 & 44.94 & 61.39 & 51.63 & 54.55 & 48.48 & 55.41 & 4.30 & 36.06 & 45.31 \\
    TimeChat-Online-7B~\cite{timechatonline} & 1 fps & 75.20 & 46.80 & 70.70 & 47.80 & 69.30 & 61.40 & 61.90 & 55.90 & 59.50 & 9.70 & 41.70 & 51.80 \\
    StreamForest-7B~\cite{zeng2025streamforestefficientonlinevideo} & 1 fps & 68.46 & 53.21 & 71.55 & 47.75 & 65.35 & 60.87 & 61.20 & 58.92 & 64.86 & 32.26 & 52.02 & 56.61 \\
    \midrule
    \multicolumn{14}{c}{\textbf{Training-free Offline-to-Online Methods}} \\
    \midrule
    LLaVA-OV-7B~\cite{li2024llavaonevisioneasyvisualtask} & 64 & 67.79 & 55.05 & 72.41 & 48.31 & 72.28 & 62.50 & 63.06 & 57.24 & 55.41 & 18.28 & 43.64 & 53.35 \\
    \hspace{3pt} + ReKV~\cite{di2025streamingvideoquestionansweringincontext} & 0.5 fps &  52.35 & 54.13 & 69.83 & 43.26 & 67.33 & 57.07 & 57.33 & 57.58&  56.08 & 18.82 & 44.16 & 50.75\\
     \rowcolor{gray!20} \hspace{3pt} + HERMES (6K tokens) & 0.5 fps & \textbf{72.48} & \textbf{62.39} & 69.83 & 47.75 & \textbf{73.27} & 64.67 & 65.07 & \textbf{61.28} & 58.78 & 26.34 & 48.80 & 56.94 \\
    
     \rowcolor{gray!40} \hspace{3pt} + HERMES (4K tokens) & 0.5 fps & \textbf{72.48} & \textbf{62.39} & \textbf{74.14} & \textbf{50.56} & \textbf{73.27} & \textbf{65.22} & \textbf{66.34} & 60.61 & \textbf{61.49} & \textbf{28.49} & \textbf{50.20} & \textbf{58.27} \\
     
    \midrule
    LLaVA-OV-0.5B~\cite{li2024llavaonevisioneasyvisualtask} & 64 & 53.69 & 53.21 & 48.28 & \textbf{33.71} & 60.40 & \textbf{48.91} & 49.70 & 46.13 & 45.27 & \textbf{12.37} & 34.59 & 42.15 \\
    \hspace{3pt} + ReKV~\cite{di2025streamingvideoquestionansweringincontext} & 0.5 fps &  41.61 & 44.95 & 50.00 & 29.78 & 60.40 & 35.87 & 43.77 & 46.13&  43.92 & 9.14 & 33.06 & 38.42 \\
     \rowcolor{gray!20} \hspace{3pt} + HERMES (6K tokens) & 0.5 fps & \textbf{57.05} & \textbf{49.54} & 55.17 & 32.58 & 60.40 & 47.28 & 50.34 & \textbf{47.81} & 47.30 & 9.14 & 34.75 & 42.55 \\
     \rowcolor{gray!40} \hspace{3pt} + HERMES (4K tokens) & 0.5 fps & 56.38 & 47.71 & \textbf{56.90} & 32.02 & \textbf{62.38} & \textbf{48.91} & \textbf{50.72} & \textbf{47.81} & \textbf{47.97} & 8.60 & \textbf{34.80} & \textbf{42.76} \\
     
    \midrule
    Qwen2.5-VL-7B~\cite{bai2025qwen25vltechnicalreport} & 1 fps & 67.79 & 55.05 & 67.24 & 42.13 & 66.34 & 60.87 & 59.90 & \textbf{51.52} & 58.78 & 23.66 & 44.65 & 52.28 \\
    \rowcolor{gray!20} \hspace{3pt} + HERMES (6K tokens) & 1 fps & \textbf{85.91} & 60.55 & \textbf{74.14} & 52.81 & 70.30 & \textbf{66.85} & 68.42 & 49.49 & 61.49 & 33.33 & 48.10 & 58.26 \\
     \rowcolor{gray!40} \hspace{3pt} + HERMES (4K tokens) & 1 fps & 85.23 & \textbf{64.22} & 71.55 & \textbf{53.37} & \textbf{74.26} & 65.22 & \textbf{68.98} & 48.48 & \textbf{62.16} & \textbf{37.63} & \textbf{49.43} & \textbf{59.21} \\
      
    \midrule
    Qwen2.5-VL-32B~\cite{bai2025qwen25vltechnicalreport} & 1 fps & 77.18 & 58.72 & 68.10 & 50.56 & \textbf{74.26} & 57.61 & 64.40 & \textbf{58.59} & 62.84 & 29.57 & 50.33 & 57.37 \\  
     \rowcolor{gray!20} \hspace{3pt} + HERMES (6K tokens) & 1 fps & 87.25 & \textbf{66.06} & \textbf{74.14} & 57.30 & 71.29 & 75.54 & 71.93 & 55.56 & \textbf{70.27} & 47.31 & \textbf{57.71} & \textbf{64.82} \\
     \rowcolor{gray!40} \hspace{3pt} + HERMES (4K tokens) & 1 fps & \textbf{88.59} & 65.14 & \textbf{74.14} & \textbf{58.99} & 71.29 & \textbf{76.09} & \textbf{72.37} & 52.19 & 66.22 & \textbf{47.85} & 55.42 & 63.90 \\

     \midrule
    Qwen3-VL-8B~\cite{bai2025qwen3vltechnicalreport} & 2 fps & 83.89 & 61.47 & 75.86 & \textbf{59.55} & 71.29 & 59.78 & 68.64 & \textbf{61.28} & 69.59 & 10.22 & 47.03 & 57.84 \\
    \rowcolor{gray!20} \hspace{3pt} + HERMES (6K tokens) & 2 fps & 85.91 & 71.56 & \textbf{81.03} & 56.74 & \textbf{72.28} & \textbf{71.74} & 73.21 & 54.88 & \textbf{70.95} & 14.52 & 46.78 & 60.00 \\
     \rowcolor{gray!40} \hspace{3pt} + HERMES (4K tokens) & 2 fps & \textbf{87.92} & \textbf{74.31} & 79.31 & 56.18 & 70.30 & \textbf{71.74} & \textbf{73.29} & 55.56 & 67.57 & \textbf{24.73} & \textbf{49.28} & \textbf{61.29} \\
      
    \midrule
    Qwen3-VL-4B~\cite{bai2025qwen3vltechnicalreport} & 2 fps & \textbf{83.89} & 66.97 & 76.72 & 57.30 & 72.28 & \textbf{66.85} & 70.67 & \textbf{60.27} & \textbf{66.22} & 23.66 & 50.05 & 60.36 \\  
     \rowcolor{gray!20} \hspace{3pt} + HERMES (6K tokens) & 2 fps & 81.88 & \textbf{73.39} & 77.59 & 58.43 & 73.27 & \textbf{66.85} & 71.90 & 54.55 & 62.84 & 44.62 & 54.00 & 62.95 \\
     \rowcolor{gray!40} \hspace{3pt} + HERMES (4K tokens) & 2 fps & 81.88 & 72.48 & \textbf{78.45} & \textbf{59.55} & \textbf{75.25} & 66.30 & \textbf{72.32} & 55.89 & 63.51 & \textbf{45.70} & \textbf{55.03} &\textbf{63.68} \\

    \bottomrule
    \end{tabular}
    \end{adjustbox}
\end{table*}
\section{Case Study}
\label{app:more_case}
\begin{figure*}[p]
  \centering
    \includegraphics[width=\linewidth]{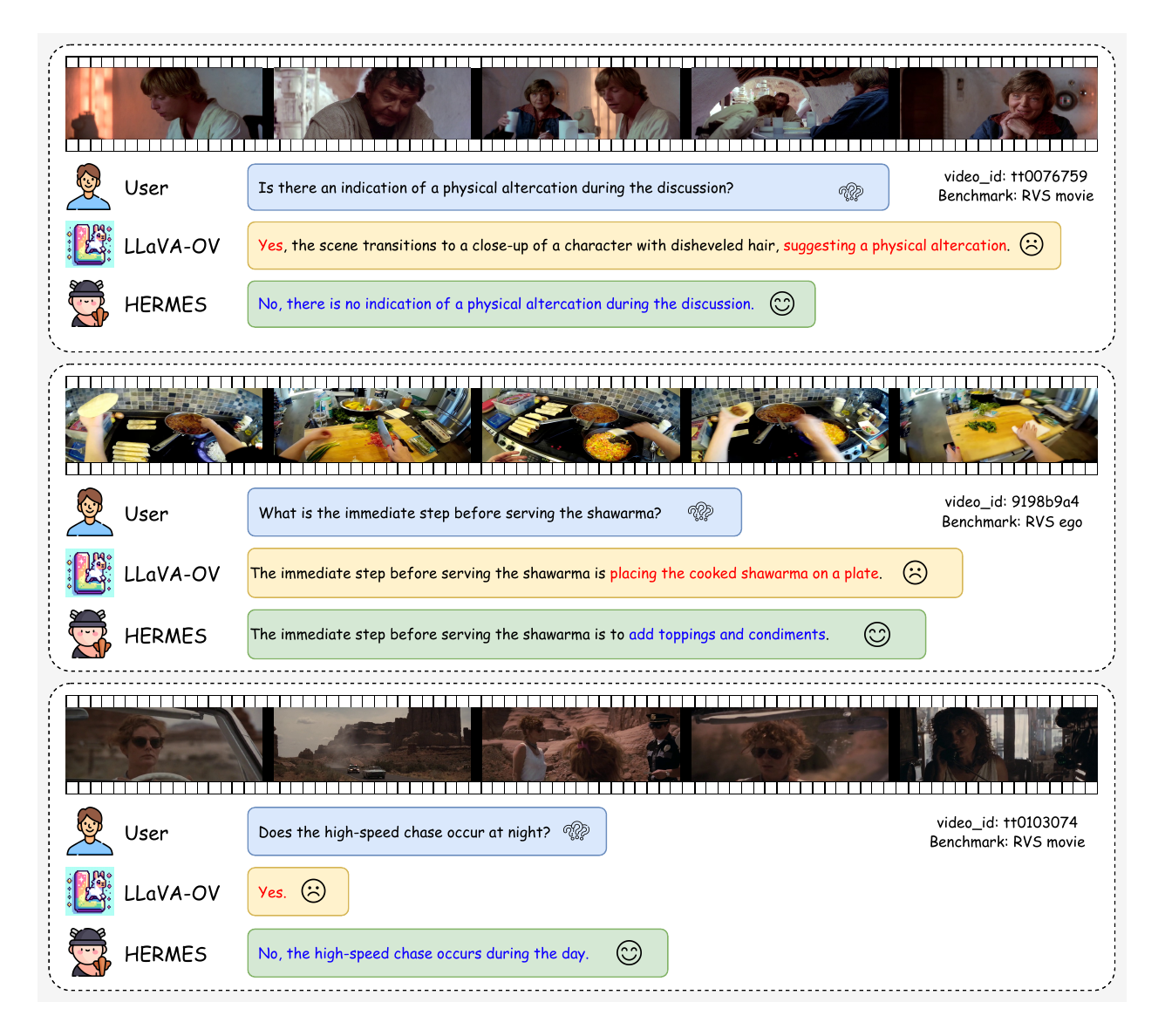}
  \caption{Cases demonstrating the superior fine-grained temporal understanding capability of \hermes relative to the \llava base model.}
  \label{fig:case_temporal}
\end{figure*}

\begin{figure*}[p]
  \centering
    \includegraphics[width=\linewidth]{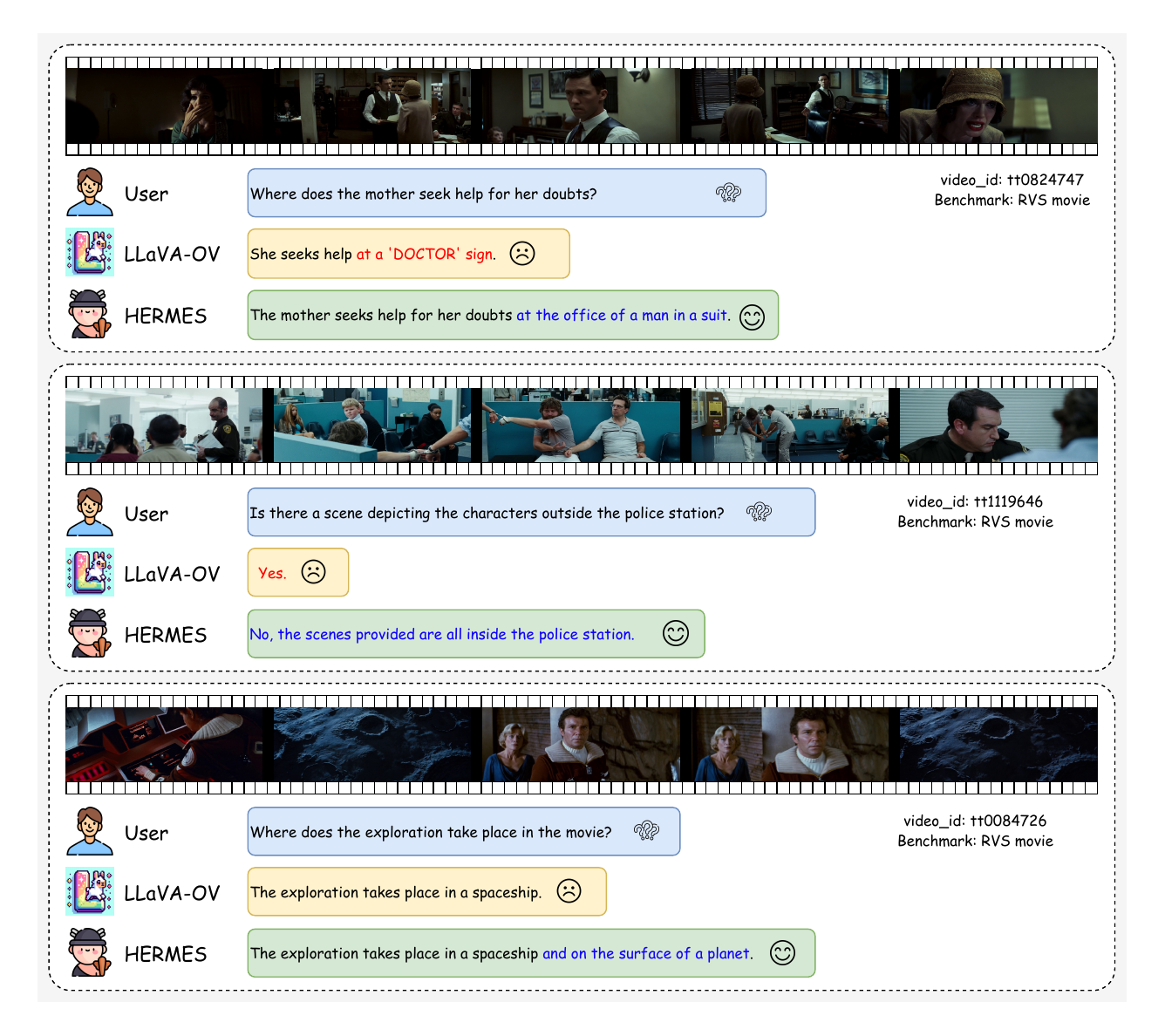}
  \caption{Cases demonstrating the superior fine-grained spatial understanding capability of \hermes relative to the \llava base model.}
  \label{fig:case_spatial}
\end{figure*}

We provide six representative case study examples from RVS-Ego and RVS-Movie to demonstrate the advantages of \hermes compared to the foundation model \llava. During the understanding of streaming long videos, \hermes exhibits significantly finer-grained temporal (shown in~\cref{fig:case_temporal}) and spatial understanding~\cref{fig:case_spatial} capabilities than its corresponding foundation model. 


\end{document}